\documentclass[11pt]{article}

\usepackage[preprint]{acl}
\usepackage{enumitem}
\setlist{nosep}
\usepackage{times}

\usepackage{amsmath}
\usepackage{amsfonts}
\usepackage{booktabs}
\usepackage{placeins}
\usepackage{subcaption}

\usepackage[T1]{fontenc}
\usepackage[utf8]{inputenc}

\usepackage{microtype}

\usepackage{inconsolata}
\usepackage[most]{tcolorbox}
\usepackage{graphicx}
\usepackage{float}

\renewcommand{\t}{\textnormal}
\title{Where A Small Language Model Helps in Invoice Categorisation, Understood Through Embedding Geometry}

\author{Emma Ceccherini \thanks{Work performed in partnership with System Holdings Limited.} \\
  University of Bristol, U.K. \\ \\\And
  Daniel Lawson \\
  University of Bristol, U.K. \\\\\And
  Anjulika Salhan \\
 System Holdings Limited \\
}

\begin{document}
\maketitle
\begin{abstract}
Categorising invoices into the correct General Ledger (GL) code underpins financial reporting and tax compliance. This is a skilled accounting judgement rather than a routine task: the correct category depends subtly on the nature of the purchasing business, the vendor and the invoice text. Whilst AI is increasingly being adopted across industries to automate tasks, including invoice categorisation, implementations built on in-house small language models (SLMs) can simultaneously reduce cost and improve data security, confidentiality, and interpretability. 
We investigate this approach by first analysing the pre-trained embedding geometry of a small sentence transformer (SBERT) and classic SLM (DeBERTa). The sentence-embedding space of this financial corpus is globally anisotropic but composed of locally isotropic clusters, extending prior token-level findings to sentence embeddings in a financial setting, and these clusters are strongly correlated with the vendor identity. 
SBERT fine-tuned on a single GPU reaches 0.96 accuracy on invoice classification, above both a zero-shot LLM and a vendor identity baseline, increasing performance for smaller, challenging categories and new clients. 
For this important generalisation problem, SBERT reaches 0.9 F1 with roughly 100 client-specific invoices, showing that an in-house SLM implementation is promising. Combining these results with geometric analysis shows that pre-trained embedding geometry is associated with classification performance and reveals a counterintuitive finding that a structured input that would help a human reader does not improve the SLM performance.

\end{abstract}

\section{Introduction}

Every invoice generated by a business needs to be assigned the correct General Ledger (GL) code. This is a nuanced accounting task that requires specialised skills, and its success underpins financial reporting and tax compliance, under UK standards (FRS 105, FRS 102 Section 1A). A reduction of errors in invoice classification benefits small and medium enterprises (SMEs) through improved financial management, reduced compliance costs and exposure to risk through penalties. 
The UK tax gap is estimated around £59 billion, for which roughly half is attributed to errors or failure to take reasonable care \cite{HMRC_tx_gap}. Hence, the motivation for this research is clear, as improving automated mapping of invoices to GL codes would improve compliance, benefiting HM Revenue and Customs (HMRC) through increased compliance and SMEs through reduced exposure to fines and compliance costs.  

AI is increasingly adopted to automate exactly these types of tasks. Large language models (LLMs) are the default choice, and some could perform well on this task. However, we find that off-the-shelf zero-shot (small) LLMs are not well suited to this specialised task (Section \ref{sec:class_frozen}).

On the other hand, in-house small language models (SLMs) are competitive in the financial domain \citep{bucher2024fine,alessandro2025categorising, garcia2020identifying} and offer concrete advantages.
They mitigate serious concerns around data security and confidentiality \citep{das2025security, wang2025unique} for a regulated domain like accounting, and significantly reduce costs \citep{belcak2025small}. Finally, SLMs improve interpretability \citep{anwar2024foundational,luo2024understanding} through a large suite of tools available for encoder-only SLMs like BERT \cite{rogers2020primer, hewitt2019structural, clark2019does}, which are not available for closed external LLMs and which may eventually be required both for audit and regulatory purposes and for public trust.

Classification performance is a critical benchmark but an examination of the internal representation of data can offer a pathway towards intrinsic interpretability \cite{doshivelez2017rigorousscienceinterpretablemachine, rudinStopExplainingBlack2019}. For the end user, interpretability promises a quantification of uncertainty \cite{Liang2023Conformal}, increased robustness \cite{geirhosShortcutLearningDeep2020}, control over associations with protected categories, and human-intuitive representation of semantic meaning \cite{Saxe2023SemanticNN}. 
These properties can be derived in part from the intermediary of geometric properties, i.e. clustering, separability and manifold structure of the neural network \cite{gao2018representation, cai2021isotropy, nastase2025testing} and can be linked to downstream behaviour \citep{rajaee2021does, hammerl2023exploring} relating to features and labels. We therefore complement classification metrics with a geometric analysis, not as an end in itself but as a foundation for developing reliable, auditable and interpretable SLMs.

\textbf{Contribution.} :
\begin{enumerate}
    \item We show \emph{where} a language model can make a valuable contribution over a vendor-only baseline, i.e. in smaller and challenging categories, which informs deployment implementations.
    
    \item We show that a small \emph{sentence transformer}, SBERT ($\sim$22M parameters) \citep{reimers-2019-sentence-bert}, fine-tuned on a single GPU, reaches 0.96 accuracy, substantiating SLMs as a potentially reliable and cost-effective route for deployment-level classification. 
    
    \item We extend prior token-level findings on globally anisotropic representation spaces with isotropic clusters to sentence embeddings; to the best of our knowledge, the first such result in a financial setting, and we show these clusters are highly associated with the vendors.

    \item We find that embedding geometry helps explain SBERT's increased performance over larger \emph{token transformer} DeBERTa ($\sim$184M) \cite{he2023debertav3improvingdebertausing}: SBERT's pre-trained space has a higher effective dimensionality and lower average cosine similarity than DeBERTa's. And DeBERTa's CLS sentence embedding collapses as the input grows richer. 

    \item We find that on unseen clients the otherwise competitive vendor baseline collapses, and SBERT improves macro F1 over DeBERTa. This suggests that line-item text in learned representations with `good geometry' carries semantic signal that transfers to unseen vendors. This is deployment-relevant as it directly measures client onboarding.
        
    \item The \emph{utility} of geometric analysis is twofold. Firstly, for this financial corpus, pre-trained embedding geometry is consistent with downstream performance and robustness. Secondly, it guides input representation: a more costly structured input useful for human reading does not improve the SLM performance due to increased similarity in the embedding space, directly relevant for practitioners.
\end{enumerate}

\textbf{Related work.} Several works have used machine learning in the accounting domain \citep{berghaus2025multi, chen2025framework, noels2024topoledgerbert, tarawneh2019invoice}. Some specifically address the classification of invoices into GL codes, including
frameworks based on static word embedding \citep{jorgensen2021machine, kotepuchai2024tree, bardelli2020automatic} or deep learning \citep{kieckbusch2020towards, munoz2022hierarchical, chi2024data, liu2021categorization}. Only a few approaches exploited modern language models \citep{munoz2022hierarchical, chi2024data}.  However, to the best of our knowledge, none of the existing works implements sentence-based SLMs and combines it with a thorough geometric investigation. 

The geometry of language models' representation spaces has been well studied \citep{ethayarajh2019contextual, cai2021isotropy}, with attempts to formalise these phenomena \citep{allen2019convergence, gao2018representation}. To the best of our knowledge, we are the first to use these geometric tools to offer insight into classification performance on a financial corpus of invoices.

\section{Methods}

\subsection{Dataset}
We analyse the problem of automatic invoice classification into GL codes. Our confidential dataset consists of real invoices from seven clients of a UK accountancy firm that span various business domains. The firm provides its clients with software that uses OCR to extract relevant textual information from a PDF or an image of the invoice; we take that OCR output as our input. Each invoice is associated with the correct accounting category, manually labelled by professional accountants. 

The full dataset comprises 2,423 invoices over 42 categories (Figure \ref{fig:categories_distribution}), 
with an average of 75 tokens per invoice. The data exhibits the severe class imbalance typical of this domain: only 14 categories contain more than 20 samples, and the largest category (\textit{Purchases}) accounts for 705 invoices. We restrict our experiments to well-populated categories: the 12 categories with more than 40 samples (2,168 invoices) defined in Table \ref{tab:account_categories}. These contain enough labelled data to train and evaluate reliably, and together they account for the majority of invoice volume. 

From the OCR output, we construct three different textual inputs, with examples in Figure~\ref{fig:invoice_examples}:
\begin{enumerate}[label=(\Alph*)]
    \item Vendor name, date, and price only, excluding the line-item descriptions.
    \item Vendor name, date, line-item descriptions, and price, as returned by the OCR.
    \item As (B) with an explicit label (\textit{Vendor Name:}, 
    \textit{Invoice Date:}, \textit{Line Items:}, \textit{Price:}).
\end{enumerate}

\begin{figure}[h]
    \centering

    \begin{tcolorbox}[colback=gray!20, colframe=gray!20, 
                      arc=4pt, boxrule=0pt]
                          \small
    \textbf{Example --- Label: Purchases} \\[2pt]
    \textit{(A)} TESCO, 12-05-2026, 4.45 \\[2pt]
    \textit{(B)} TESCO, 12-05-2026, 75892 Red Pepper 1 0.70 
    27836 Chocolate Cake PLT PION 390g 3.75, 4.45 \\[2pt]
    \textit{(C)} Vendor Name: TESCO, Invoice Date: 12-05-2026, 
    Line Items: 75892 Red Pepper 1 0.70 27836 Chocolate Cake 
    PLT PION 390g 3.75, Price: 4.45 

    \end{tcolorbox}

    \caption{Fictitious example of the three input variants.}
    \label{fig:invoice_examples}
\end{figure}

\subsection{Experimental set up}

We use the sentence transformer \textbf{{SBERT}} \citep{reimers-2019-sentence-bert} (\emph{all-MiniLM-L6-v2} ), an encoder-only transformer designed and trained for sentence embedding with $\approx$22M parameters and 384 model dimension. Each invoice input is embedded from the last layer extracted with SBERT native pooling. 
We compare it to \textbf{DeBERTa} (\emph{microsoft/deberta-v3-base}) with $\approx$184M parameters and 768 model dimensions. {DeBERTa} computes token embeddings which must be pooled to compute sentence embeddings. We consider the two most popular solutions: pooling using the `CLaSsification' token embedding [CLS], vs pooling using the mean embedding of all tokens [mean], both in the last layer. The code can be found in this \href{https://github.com/emmaceccherini/SH-SLM-code}{repository}.

\section{Geometric analysis}\label{sec:geometry}
We analyse the pretrained embedding geometry along four 
diagnostics: effective dimensionality, global anisotropy, 
unsupervised clustering structure, and local isotropy.

\begin{figure*}[t]
    \centering
    \includegraphics[width=0.95\linewidth]{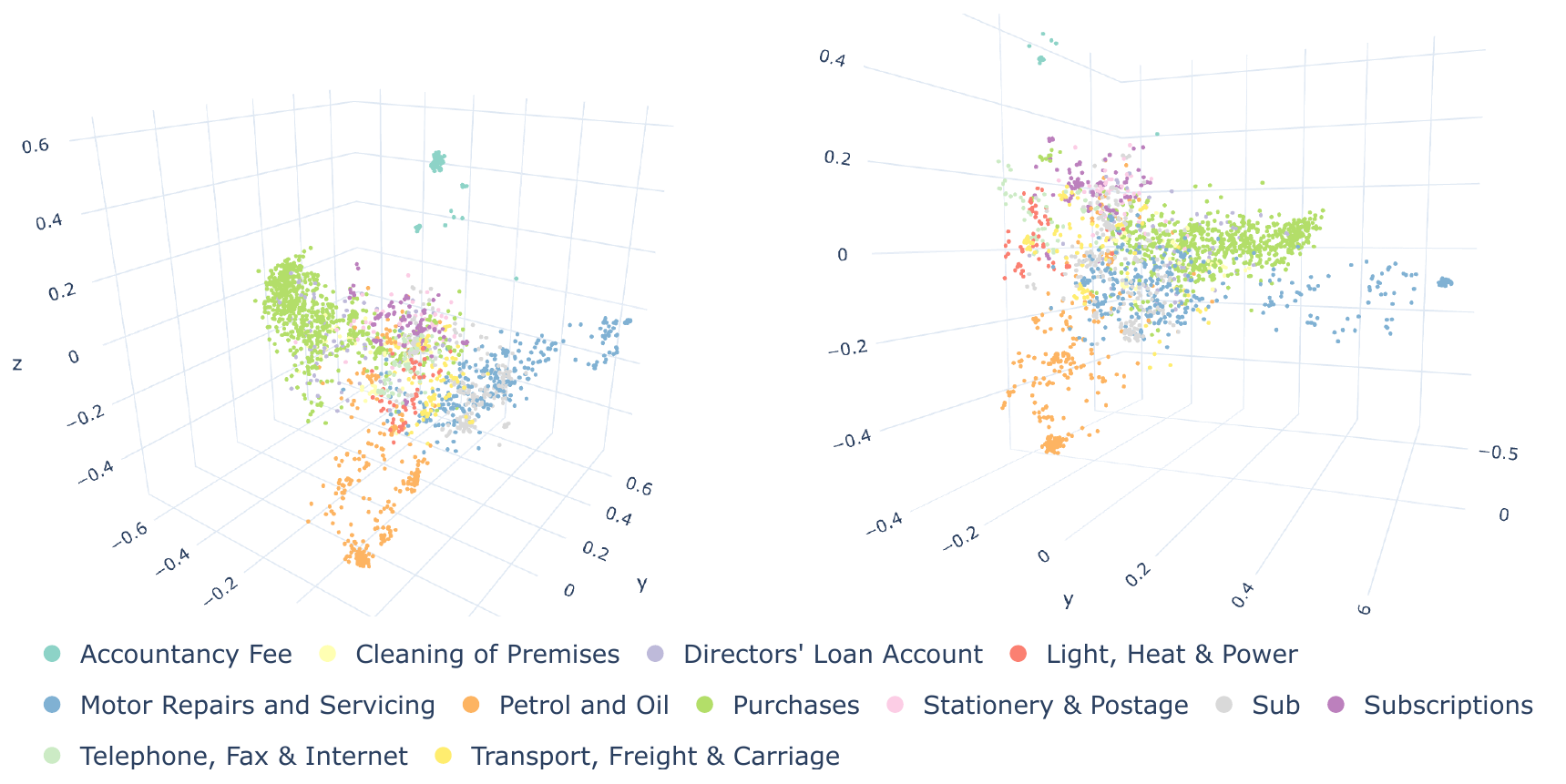}
    \caption{ {\bf Frozen transformer} visualisation of the first three principal components of the representation space of pre-trained SBERT with input B. Note the overlap between Purchases (green) and Directors Loan Account (light purple). }
    \label{fig:pca_base}
\end{figure*}

\subsection{Effective dimensionality}
Inspired by \citet{cai2021isotropy}, we first look at the effective dimensionality of the embedding spaces, via centred PCA. For each combination of model and input type, we start with the embedding matrix $M \in \mathbb{R}^{n\times m}$, where n = 2,168  and m = 384 for SBERT and 768 for DeBERTa. Let 
$\sigma_1 \geq \sigma_2 \geq \cdots \geq \sigma_m$ be the 
eigenvalues of $M$'s covariance matrix, and define the 
explained variance ratio $r_k = \sum_{i=1}^{k} \sigma_i 
\,/\, \sum_{i=1}^{m} \sigma_i$ and the $\epsilon$-effective dimension to be $d(\epsilon) := \t{argmin}_k r_k \geq \epsilon$. For each model and input type, we report $d(0.8)$, the number of dimensions necessary to explain 80\% of the variance in the embedding space, in Table \ref{tab:effective_dim}.

\begin{table}[h]
    \centering
    {\small
    \begin{tabular}{lcccc}
        \hline
        Model & A & B & C &$d_{model}$\\
        \hline
        SBERT        & 42 & 61 & 57 & 384\\
        DeBERTa mean & 13 &  7 &  6 & 768\\
        DeBERTa CLS &  10 &  1 &  1 & 768\\
        \hline
    \end{tabular}}
        \caption{Effective dimensionality (80\% variance explained) by model and input type.}
    \label{tab:effective_dim}
\end{table}

SBERT has higher effective dimensionality across all input types, as it's trained to produce semantically rich embeddings. SBERT embeddings for input types B and C have a higher $d(0.8)$ than A, reflecting the additional textual content, with little difference between B and C themselves. 

Figure \ref{fig:pca_base} shows the first 3 principal components of SBERT (B) embeddings. The embedding of \textit{Accountancy fee} forms a tight cluster. Since every client in our dataset uses the same UK accountancy firm, these invoices are highly similar to each other. The rest of the data is distributed along an X-shaped manifold. The three biggest branches correspond to the three largest categories \textit{Purchases} (green), \textit{Motor Repairs and Servicing} (blue) and \textit{Petrol and Oil} (orange). The fourth, smaller, branch is made up of \textit{Light, Heat $\&$ Power} (red) \textit{Subscriptions} (purple), \textit{Telephone, Fax $\&$ Internet} (light green). Notably, SBERT, without any fine-tuning, picked up the semantic similarity that groups these last three categories as `recurring bills'. Figure \ref{fig:PCA_color_vendors} shows the association with Vendor ID.

DeBERTa has a larger embedding dimension with low effective dimension, especially for CLS pooling. This reflects the well-known fact that CLS embeddings tend to have semantically `meaningless' dominant directions \citep{liang2021learning}; here, the first PC is highly correlated ($>$ 90\%) with the norm of the embeddings for all input types. 

\subsection{Anisotropy}\label{sec:anistropy}

Extensive literature has demonstrated that, counterintuitively, well-performing SLMs and LLMs do not use all of the embedding space; this is called anisotropy \citep{cai2021isotropy, gao2018representation, nastase2025testing}. We measure this globally using the cosine similarity between any two pairs of embeddings in Table~\ref{tab:cosine} which quantifies the degree to which the embeddings exist along shared directions.

\begin{table}[h]
    \centering
       { \small
    \begin{tabular}{lccc}
        \hline
        Model & A & B & C \\
        \hline
        SBERT        & 0.365 & 0.282  & 0.496\\
        DeBERTa mean & 0.973  &0.933 & 0.949\\
        DeBERTa cls &  0.999 &  0.992& 0.993\\
        \hline
    \end{tabular}}
        \caption{\textbf{Frozen transformers} average cosine similarity between any embedding pairs by model and input type.}
    \label{tab:cosine}
\end{table}

Consistent with the effective dimensionality results, both 
DeBERTa pooling strategies produce extremely high average 
cosine similarity, i.e. these sentence embeddings live in a 
narrow cone of the representation space. SBERT embeddings have substantially lower average cosine similarity, i.e., they are more homogeneously distributed, though still anisotropic (cosine similarity 0.3-0.5).

To analyse class-level structure, let $C_k$ denote the 
set of indices in class $k$, and define the within-class and between-class cosine similarity:

\[
\overline{s}_w = \frac{1}{K} \sum_{k=1}^{K} 
\frac{1}{|C_k|(|C_k|-1)} 
\!\!\sum_{\substack{i,j \in C_k, \\ i \neq j}}\!\! 
\cos(M_i, M_j),
\]
\[
\overline{s}_b = \frac{1}{K} \sum_{k=1}^{K} 
\frac{1}{|C_k|\,(n - |C_k|)} 
\sum_{\substack{i \in C_k,\\ j \notin C_k}} \cos(M_i, M_j).
\]

Table \ref{tab:cosine_wb_combined} reports  SBERT and DeBERTa-mean $\overline{s}_w$ and  $\overline{s}_b$ with selected models, cluster definitions and origin; Table~\ref{tab:cosine_wb_true_full} shows every pairing, with example distributions in Figure \ref{fig:cosine_grid}.

As expected, the cosine similarity within classes is higher than that between classes (True labels, uncentered). 
Notably, Input C has both larger within and between cosine similarity than B (and A). This is not an artefact: the input texts in dataset C are genuinely more similar to one another, because the added field labels (\textit{Vendor name}, \textit{Price}, etc.) introduce seven identical words for each document. Unintuitively, Input C, the structured input, which would be helpful for a human reader, makes the inputs harder for a language model to distinguish. 

\begin{table}
\centering
{\small
\begin{tabular}{l|c|c|c}
\hline
Model & \multicolumn{1}{c|}{A} & \multicolumn{1}{c|}{B} & \multicolumn{1}{c}{C} \\
\hline
SBERT         & 54
& 59 
& 60
\\
DeBERTa mean & 52 
&  2 
&  2 
\\
DeBERTa CLS  & 52 
&  2 
&  2 
\\
\hline
\end{tabular}}
\caption{{\bf Frozen transformers} number of $K$-means clusters $|G|$ selected by highest silhouette score, by model and input type.}\label{tab:clustering}
\end{table}

\subsection{Clustering}\label{sec:clustering}
\begin{table*}[t]
\centering
{\small
\setlength{\tabcolsep}{4pt}
\begin{tabular}{ll cccc cccc}
\toprule
& & \multicolumn{4}{c}{True labels} & \multicolumn{4}{c}{K-means clustering} \\
\cmidrule(lr){3-6} \cmidrule(lr){7-10}
& & \multicolumn{2}{c}{Uncentered} & \multicolumn{2}{c}{Centred}
   & \multicolumn{2}{c}{Uncentered} & \multicolumn{2}{c}{Centred} \\
\cmidrule(lr){3-4} \cmidrule(lr){5-6} \cmidrule(lr){7-8} \cmidrule(lr){9-10}
Type & Model & Within & Between & Within & Between
            & Within & Between & Within & Between \\
\midrule
B & SBERT        & 0.478 & 0.249 & 0.003 & 0.000 & 0.708 & 0.265 & -0.015 & 0.000 \\
B & DeBERTa mean & 0.938 & 0.930 & 0.013 & 0.009 & 0.942 & 0.841 &  0.002 & 0.001 \\
\midrule
C & SBERT        & 0.641 & 0.485 & 0.003 & 0.000 & 0.823 & 0.476 & -0.031 & 0.000 \\
C & DeBERTa mean & 0.956 & 0.948 & 0.019 & 0.011 & 0.953 & 0.864 &  0.000 & 0.001 \\
\bottomrule
\end{tabular}}
\caption{{\bf Frozen transformer} average within-class/cluster and between-class/cluster cosine similarity,
using the true labels and the K-means clustering partition.}
\label{tab:cosine_wb_combined}
\end{table*}
We now repeat the analysis of Section \ref{sec:anistropy} with unsupervised clustering, instead of true labels. We perform K-means clustering of the embeddings and report the number of clusters $|G|$ in Table \ref{tab:clustering} which yields the `most separable clusters' as measured by mean silhouette score for $G \in [1,60]$

The DeBERTa-CLS clustering on inputs B and C collapses to two tight (silhouette $> 0.6$) clusters, and cluster membership is highly correlated with embedding norm: the two clusters simply separate `large' and `small' embeddings. We discuss the collapse in more detail in Section \ref{sec:cls_collapse}.

SBERT embeddings are partitioned into a similar number of clusters for all input types, with tighter clusters for input A (silhouette score 0.5 vs.\ $\sim$0.3 for B and C). This clustering is strongly aligned with vendor identity: normalised mutual information (NMI) between vendor identity and cluster membership is $>0.8$ across input types.  NMI with true labels is lower (0.5-0.6), indicating some alignment between the clustering and the true labels. The clustering of input A is the most aligned with vendor identity as expected, since the vendor name is the only textual signal in input A. Inputs B and C clusterings are slightly more label-aligned, suggesting that adding the line-item text spreads the embeddings along directions aligned with the true accounting categories.

In the best-case scenario, SBERT with input B (SBERT-B), the clustering accounts for 61\% of the uncertainty in the labels. Combined with the lowest average cosine similarity and the highest effective dimensionality (among SBERT variants), this shows that SBERT-B displays the best geometric properties, which we later show aligns with the classification results in Section \ref{sec:classification}.

Table \ref{tab:cosine_wb_combined} - clustering, uncentered - (full Table \ref{tab:cosine_wb_cluster_full}) reports the within- and between-cluster cosine similarity. Compared to the results with true labels, the within-cluster similarity is higher, while the between-cluster similarity is similar. This supports the hypothesis that the pre-trained embeddings identify the vendor name partition more sharply than the accounting category.

\subsection{Local isotropy}

Following \citet{cai2021isotropy}, we recompute the 
within- and between-class (or cluster)  after subtracting the class (or cluster) mean. In this \emph{local} sense, the sentence embeddings are almost perfectly isotropic (Table \ref{tab:cosine_wb_combined}).

\citet{cai2021isotropy} found local isotropy for \emph{token} embedding. Our result extends theirs to \emph{sentence} embeddings of invoices: the embedding space is a globally anisotropic manifold composed of locally isotropic clusters.

We further observe that centring by true labels and centring by $K$-means clusters recover isotropy to a similar degree, suggesting that both partitions are, in a different way but to an equal degree, the cause of the anisotropy. This can be explained by the near-perfect many-to-one mapping between vendors and accounting labels.  Businesses repeatedly purchase similar goods or services from each vendor, so the vendor partition and the label partition expose nearly the same directions of variation in the embedding space.

\subsection{CLS collapse}\label{sec:cls_collapse}

DeBERTa CLS embeddings show a striking pattern across all diagnostics. Moving from input type A to B/C, the best number of clusters drops from 52 to 2, the effective dimensionality goes from 10 to 1, and the cosine similarity is $>0.99$. The CLS embeddings exhibit a consistent and severe collapse under richer textual inputs (B/C). The first principal component and the clustering partition are highly correlated with the embedding norm: the CLS embeddings recover the norm rather than any semantic property \citep{liang2021learning}.

A likely explanation is that the CLS token, unlike content tokens, attends nearly uniformly to all the other tokens in the sentence. As input length grows from A to B and C, the CLS representation is the result of iterated weighted averages over many high-dimensional embedding vectors, which completely dilutes the signal. Indeed, the sentence embedding constructed with mean pooling exhibits a less extreme but similar pattern. In the latter, the average is computed only once at the final layer rather than within every layer. 

This collapse is consistent with the DeBERTa-CLS classifier degenerating to predicting only the larger classes (Section \ref{sec:classification}) while mean pooling retains some discriminative ability but still underperforms SBERT.

\section{Classification}\label{sec:classification}

\begin{table*}[t]
\centering

\setlength{\tabcolsep}{1.6pt}

{\small
\begin{tabular}{l ccc ccc ccc c c }
\toprule
 & \multicolumn{3}{c}{{DeBERTa CLS}} & \multicolumn{3}{c}{{DeBERTa mean}} & \multicolumn{3}{c}{{SBERT}} & Vendor & LLM\\
\cmidrule(lr){2-4} \cmidrule(lr){5-7} \cmidrule(lr){8-10}
& {A} &{B} & {C} &{A} &{B} & {C} & {A} & {B} & {C}  &baseline & baseline\\
\midrule
Acc. & $0.39_{ \pm 0.02}$ &$0.62_{ \pm 0.02}$& $0.45_{\pm 0.04}$&$0.51_{\pm0.02}$ &$0.63_{\pm0.02}$&$0.62_{\pm 0.02}$& $0.70_{\pm0.04}$&$\underline{0.87}_{\pm  0.01}$&$0.83_{\pm0.02} $ &$\textbf{0.89}_{\pm 0.01}$ & $0.52_{\pm0.02}$\\
M F1  &$0.18_{\pm0.08} $&$0.55_{\pm0.02}$&$ 0.33_{\pm0.06}$&$0.39_{\pm0.03}$&$ 0.53_{\pm0.04}$&$0.49_{\pm0.02}$& $0.67_{\pm0.06}$&$\underline{0.80}_{\pm0.01}$&$0.77_{\pm0.02}$ &$\textbf{0.87}_{\pm 0.01}$ & $0.39_{\pm 0.02}$\\
W F1& $0.31_{\pm0.05}$&$0.62_{\pm0.02}$&$0.44_{\pm0.04}$&$0.49_{\pm0.01}$&$0.63_{\pm0.01}$&$0.62_{\pm0.02}$&$0.69_{\pm0.04}$&$\underline{0.85}_{\pm0.01}$&$0.82_{\pm0.02}$ &$\textbf{0.89}_{\pm0.01}$ & $0.52_{\pm 0.02}$\\

\bottomrule
\end{tabular}}
\caption{{\bf Frozen Transformers} accuracy, macro F1 and weighted F1 for classification with frozen transformers for
each model and input type. The LLM baseline is zero-shot, and the transformers use a trained linear head. We report the mean$\pm$s.d. over 5 splits. The best performance is in \textbf{bold}, the second best is \underline{underlined}.}
\label{tab:classification}
\end{table*}

\subsection{Frozen transformers
 and vendor baseline}\label{sec:class_frozen}

For each model and input type, we freeze the encoder and train a linear classification head on top of the sentence embedding; for full implementation see \ref{sec:exp_details}. In Section \ref{sec:geometry} we found that there was a strong association between the vendor identity and the labels, as businesses routinely repeat orders from the same suppliers. So we compare the language models to a parameter-free baseline that maps each vendor to its most frequent GL code in the training dataset and unseen vendors to the majority class. We also compare with a zero-shot open weight LLM  \emph{Qwen3-4B} baseline, for full implementation see \ref{sec:exp_details}.

Table \ref{tab:classification} reports the results. Without any encoder training, the vendor mapping baseline outperforms both SBERT and DeBERTa and the LLM baseline on any input type. Although the pre-trained embedding space of both models is highly correlated with vendor identity, the linear head is not enough to learn the map. The vendor-to-label map is many-to-one: with hundreds of vendors grouped in 12 classes. While a single memorised vendor is enough for the baseline to classify a small category correctly, a linear head requires more samples to learn a reliable linear decision boundary. 

Focusing on the transformers' performance, we find that the geometric properties of Section \ref{sec:geometry}  
are associated with classification performance, with SBERT on input B substantially surpassing DeBERTa mean and CLS, respectively, and input B being the best across models. Unintuitively, the structured input (C), which would help a human learner, doesn't improve the performance of a language model, due to the artificially increased within-corpus similarity described in Section \ref{sec:anistropy}. The full classification reports can be found in
Tables \ref{tab:report_SBERTB}-\ref{tab:class_rep_baseline}. The classification layer trained on DeBERTa CLS embeddings ignores ($\sim$0\% accuracy) some of the smaller categories, consistent with the collapse highlighted in Section \ref{sec:cls_collapse}.  
SBERT on input B achieves F1 scores around $0.8$ on all categories but one, regardless of the category size, showing that there is a genuine signal extractable from the embedding space, but not linearly accessible.

\subsection{Fine Tuning}\label{sec:finetuning}
\begin{table}[h]
\centering
\setlength{\tabcolsep}{3.1pt}
{\small
\begin{tabular}{lccc}
\hline
\textbf{Model}  & \textbf{Acc.} & \textbf{M F1} & \textbf{W F1} \\
\hline
SBERT 2l & $0.945_{\pm0.00}$ & $0.915_{\pm0.01}$ &  $0.944_{\pm0.00}$\\
SBERT 4l & $0.950_{\pm 0.01}$ &  $0.924_{\pm 0.02}$ & $0.950_{\pm0.01}$\\
SBERT 6l & $0.950_{\pm0.01}$ & ${0.926}_{\pm0.01}$ & $0.949_{\pm0.01}$\\
\hline
SBERT 2l b.n. & $\mathbf{0.961}_{\pm0.00}$ & $\mathbf{0.939}_{\pm0.01}$ &$0.960_{\pm0.00}$\\
SBERT 4l b.n.&$\mathbf{0.961}_{\pm0.00}$& ${0.938}_{\pm0.02}$ &$\mathbf{0.961}_{\pm0.01}$\\
SBERT 6l b.n.& $\mathbf{0.961}_{\pm0.01}$ & $0.936_{\pm0.02}$ & $0.960_{\pm0.01}$\\
\hline
DeB. CLS 2l & $0.901_{\pm 0.02}$ & $0.855_{\pm0.04}$ & $0.901_{\pm0.02}$ \\
DeB. CLS 4l& $0.381_{\pm0.05}$ & $0.149_{\pm0.04}$ & $0.323_{\pm0.04}$\\
DeB. CLS 6l&  $0.352_{\pm0.10}$ & $0.131_{\pm0.13}$ &$0.207_{\pm0.04}$\\
\hline
DeB. CLS 2l b.n. & $0.913_{\pm0.02}$ & $0.870_{\pm 0.03}$ &  $0.915_{\pm0.02}$ \\
DeB. CLS 4l b.n. &  $0.457_{\pm0.19}$ & $0.210_{\pm0.25}$ & $0.382_{\pm0.24}$ \\
DeB. CLS 6l b.n.&  $0.344_{\pm0.03}$ &  $0.077_{\pm0.04}$&$ 0.256_{\pm0.14}$\\
\hline
DeB. m 2l& $0.930_{\pm 0.01}$ & $0.894_{\pm0.02}$ & $0.930_{\pm0.01}$ \\
DeB. m 4l& $0.779_{\pm0.10}$ &$ 0.618_{\pm0.19}$ & $0.757_{\pm0.13}$ \\
DeB. m 6l&  $0.366_{\pm0.06}$ &  $0.192_{\pm0.14}$ &$0.302_{\pm0.14}$\\
\hline
DeB. m 2l b.n.& $0.950_{\pm0.01}$ &${0.916_{\pm0.02}} $& $0.950_{\pm0.01}$ \\
DeB. m 4l b.n.& $0.818_{\pm 0.04}$ & $0.673_{\pm0.11}$ & $0.807_{\pm0.05}$\\
DeB. m 6l b.n.& $ 0.480_{\pm0.14}$ & $0.245_{\pm0.04}$ & $0.389_{\pm 0.12}$ \\
\hline
\end{tabular}}
\caption{{\bf Post fine-tuning} accuracy, macro F1 and weighted F1 across fine-tuning configurations and number of unfrozen final layers (l) for input type B with/without business-nature prefix (+b.n.) reporting mean$\pm$s.d. over 5 splits. The best performance is in \textbf{bold}. }
\label{tab:ft_results_B}
\end{table}

We fine-tune both language models by training the last 2, 4 or 6 layers with the classification head, both with and without adding the business-name prefix,  e.g. `Fish and Chip restaurant' or `Logistics and Transport', to the inputs (+ b.n.). See \ref{sec:exp_details} for full implementation details. Results for input B are reported in Table \ref{tab:ft_results_B} (Figure \ref{fig:confusion_matrix} shows the confusion matrix), and full classification reports for all inputs in Table \ref{tab:class_ft_full}-\ref{tab:class_un3}. 

Fine-tuning substantially improves the performance over the head-only baselines, closing the gap and surpassing the vendor baseline. 

Fine-tuning SBERT substantially improves performance in the \textit{Director's loan account} \textit{(DLA)} category. \textit{DLA} and \textit{Purchases} are financially different categories for which invoices may be identical. Loosely, the difference does not lie in the goods or services bought, but in whether the director or the business paid, and whether the purchase is in furtherance of the business. The per-class F1 reaches 0.53 (6 layers no b.n.) and 0.71 adding the business name prefix (2 layers + b.n.), an improvement of 27 percentage points over vendor baseline. Similarly, fine-tuned SBERT improves by 10\% on the vendor baseline on the \textit{Stationery \& Postage} category. In Section \ref{sec:class_frozen}, we showed that vendor name does account for a large part of the variation in invoices, but the rich text description is essential on the long tail of `hard invoices', which is significant in a domain with inherent high-class imbalance.  

Despite being far larger, the zero-shot LLM (Qwen3-4B) baseline underperformed SBERT both with frozen encoder and especially after fine-tuning ($+0.55$ M F1). This is consistent with the literature \citep{bucher2024fine}, and corroborates our hypothesis that an in-house fine-tuned SLM is preferable to an off-the-shelf LLM in this computationally constrained setting.

DeBERTa is less robust, fine-tuning successfully only at depth 2.  
This is most likely due to the large number of DeBERTa parameters relative to our sample size. Similarly to the geometric properties, DeBERTa with mean pooling shows the same qualitative pattern as DeBERTa CLS  but less extreme, and it still underperforms SBERT.

\begin{figure*}[t]
    \centering
    \includegraphics[width=0.95\linewidth]{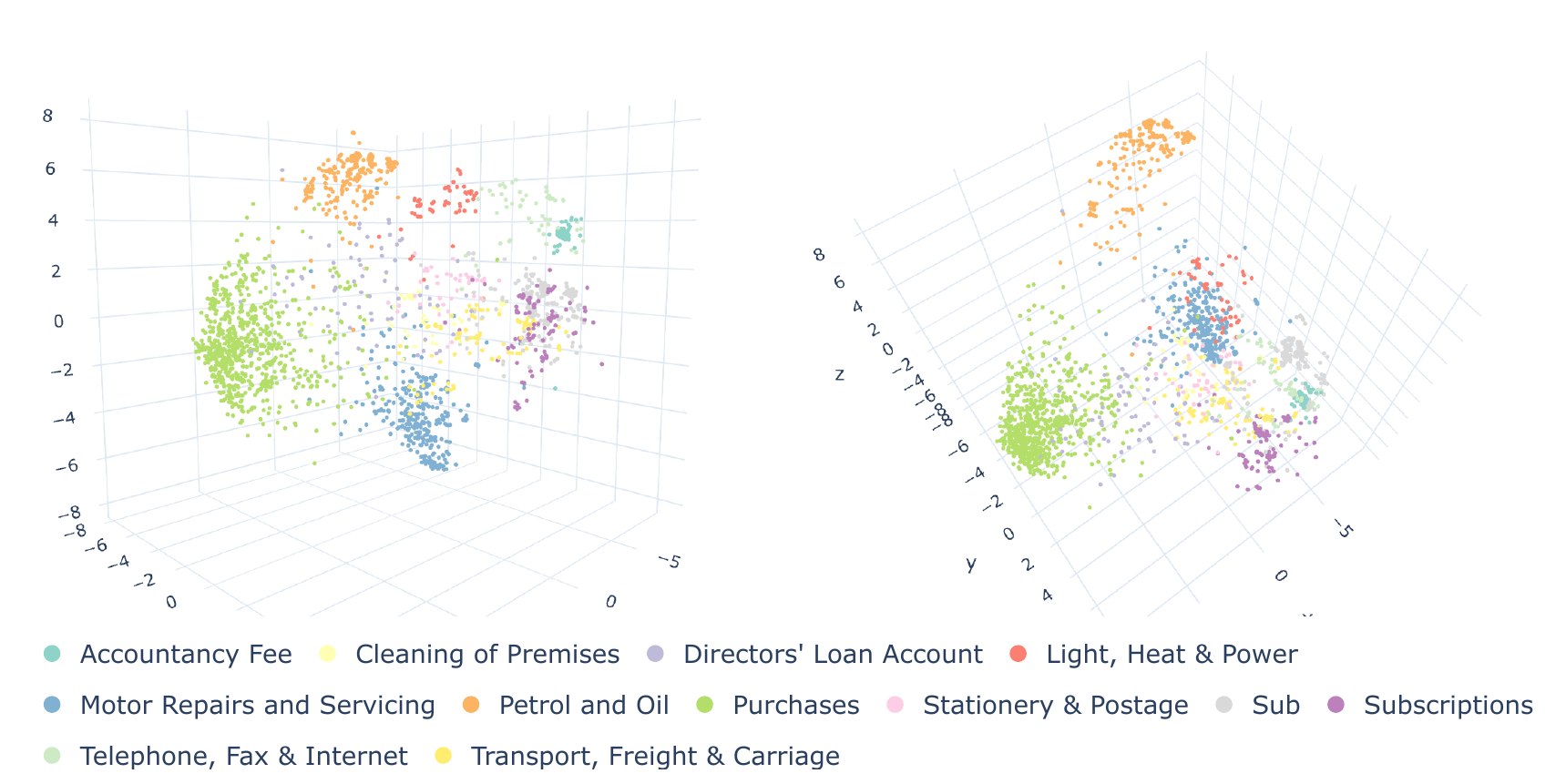}
    \caption{{\bf Post fine-tuning} visualisation of the first three principal components of the representation space of SBERT with input B. Compared to Figure \ref{fig:pca_base}, note that classes appear more distributed over the sphere, with
    Purchases (green) in the same direction but separated from Directors Loan Accounts (light purple). }
    \label{fig:pca_ft}
\end{figure*}

Figure \ref{fig:pca_ft} shows the first three principal components of the SBERT embeddings after fine-tuning the last two layers. 
Compared to the X shaped manifold of Figure \ref{fig:pca_base}, fine-tuning has `unravelled' the manifold that now qualitatively resembles a simplex with vertices at \textit{Petrol $\&$ Oil} (orange), \textit{Purchases} (green), \textit{Motor Repairs and Servicing} (blue), and the `recurring bills' which includes for example \textit{Accountancy Fee}, \textit{Subscriptions} and  \textit{Telephone, Fax $\&$ Internet}. \textit{DLA}, which is semantically close to \textit{Purchases}, is now much better separated from it, consistent with the large per-class F1 improvement. 
A simplex geometry is considered to be `optimal' for classification \cite{Weinan2022SimplexNeuralNetworks} because the classes fall on the vertices and irregularities pull points towards the centre, opening a pathway to uncertainty calibration \cite{Kull2019TemperatureScaling}, which we leave for future work.

\subsection{Generalise to new clients}

In Section \ref{sec:finetuning}, we stratified the training and testing data per category due to high class imbalance, so a given client's data likely appeared in both sets. We now investigate performance for a previously unseen client. For practitioners, this simulates onboarding a new client.

We leave one client, the `test client', out of our dataset at a time and split the remaining data 80/20 into training and validation sets, respectively. Then we repeat the experiment, injecting a fraction $\epsilon$ of the `test client' data in the train/val set (and removing it from the test set). We fine-tune from scratch the best configurations of SBERT and DeBERTa from Section~\ref{sec:finetuning} for $\epsilon \in [0, 0.8]$.

\begin{figure}[htb]
    \centering

    \includegraphics[width=0.99\columnwidth]{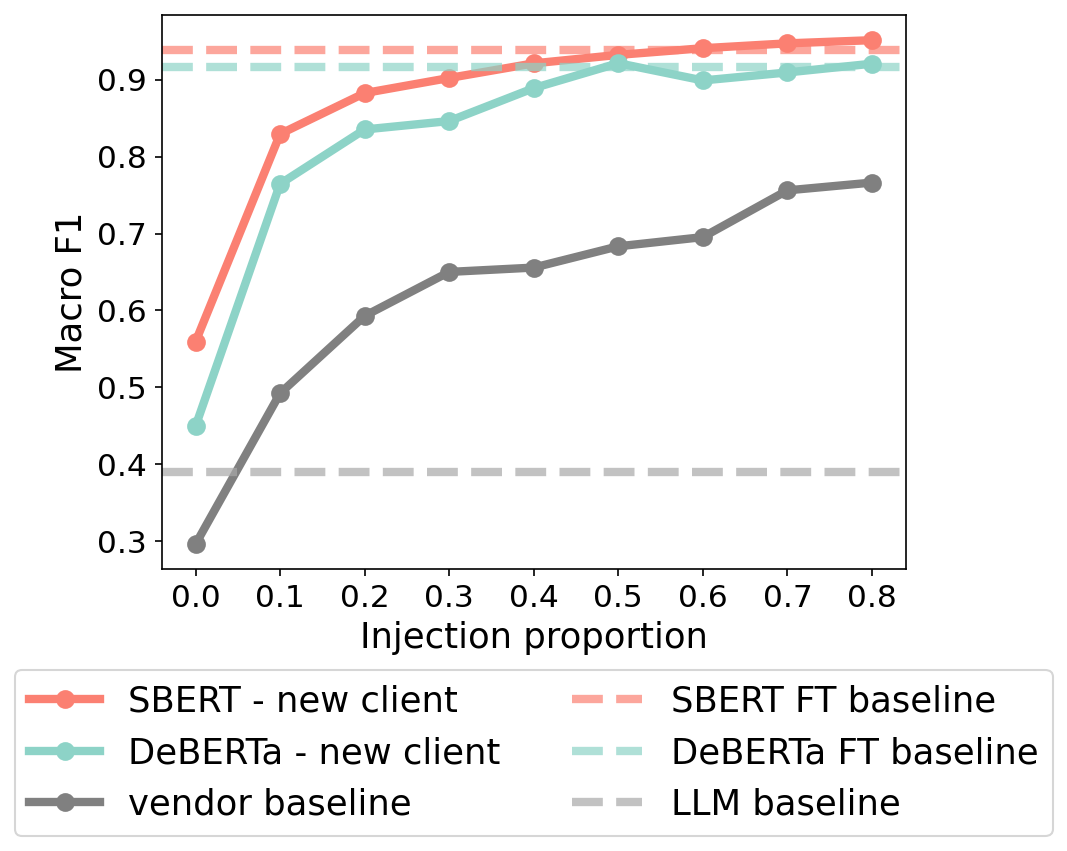}
    \caption{{\bf Left-out client experiment}, Macro F1 for fine-tuned SBERT and DeBERTa as a function of the proportion of `new client' data injected into the training set (solid line). The baselines (dashed lines) are from Section \ref{sec:finetuning}.
}
    \label{fig:gen_ex}
\end{figure}

Figure \ref{fig:gen_ex} reports the average macro F1 score on the `testing client' as a function of $\epsilon$. For a new client, the vendor baseline, which was competitive in-distribution, severely underperforms both models, since performance depends entirely on the overlap between the new client's suppliers and those seen in training. Between  SBERT and DeBERTa, the former reaches higher F1 across all values of $\epsilon$: at $\epsilon=0$, corresponding to a fully unseen client, SBERT achieves an F1 10\% higher than DeBERTa. This shows that the value of learned representations relies on their generalisation abilities: line-item text carries semantic signal that transfers to unseen vendors. The gap between SBERT and DeBERTa highlights the extent to which that signal is more easily recoverable from semantically meaningful representations with better geometry.

Both models need a warm-up period before
reaching reliable performance; to achieve an F1 of 0.9 or higher, SBERT needs, on average, 100 client-specific data points, whereas DeBERTa needs $\sim$75\% more client-specific data (174 data points). Beyond the warm-up region, both SBERT and DeBERTa continue to improve with $\epsilon$ and converge towards their fine-tuned baselines obtained in Section \ref{sec:finetuning} (dashed lines in Figure \ref{fig:gen_ex}).

This pattern is consistent with the geometric analysis and the overall findings of this paper. SBERT's higher-dimensional, less anisotropic space utilises more independent directions where a client's invoices can be separated from a few labelled examples. The generalisation gap at $\epsilon =0$ and the faster warm-up confirm that SBERT is the more robust choice in our setting, corroborating that pre-trained embedding geometry is linked to fine-tuned accuracy and robustness.

\section{Conclusion}

Whilst this experiment uses a small dataset covering a subset of all existing categories, it is a promising result for an in-house deployment where cost, data confidentiality, and interpretability matter. 
We explored geometric properties which demonstrated that previous token-level findings, that embeddings are globally anisotropic with locally isotropic clusters, apply to the sentence embedding of invoices.

While exploiting vendor identity alone reaches respectable performance, SLM became truly useful once fine-tuned, especially on smaller categories and for newly onboarded clients. This improvement when using full invoice content implies that `line item' text has been at least partially mapped to the label through semantic meaning. Moreover, whilst both models achieved similar high accuracy under naive comparison, sentence embedding produced more robust embedding that generalised better than token embeddings, consistent with the hypothesis that geometry relates to both performance and robustness.

Fine-tuned performance on the structured Input C is comparable to Input B, which is particularly relevant for practitioners. Since the structured input remains uninformative after fine-tuning, we conclude that omitting labelling of fields in the free text OCR is sufficient.  This removes an additional human-in-the-loop step, resulting in a reduction in cost and complexity. Input B requires only a single human-in-the-loop at the end, which is irreducible to comply with regulations.

These results suggest a promising direction for future research towards an uncertainty-aware model to identify the data that is confidently assigned with interpretable reasons, focusing expert attention where it is most needed.

\section*{Limitations}

Our study is based on a single confidential dataset from one UK accountancy firm, comprising 2,168 invoices across 12 categories from seven clients. Real, labelled invoice data of this kind is difficult to obtain, and this corpus reflects the access we were able to secure; we acknowledge that the sample is modest. Our results should be seen as evidence that an in-house SLM pipeline is viable in this setting, but they may not transfer directly to firms operating with different accounting conventions, in different domains or countries.

Our geometric analysis, and its connection to classification performance, is based on a comparison of two model families. To further explore this connection and make an explanatory claim, more model types and additional experiments should be analysed. We used a single set of standard training parameters for both SBERT and DeBERTa for a consistent comparison; however, this study doesn't include an exhaustive search of the hyperparameter space. 

\section*{Ethical considerations}

The data analysed in this paper is confidential: it is commercially sensitive and may contain identifying information about businesses and their suppliers. The data has been used under the firm's data-handling agreements, and it will not be released. Where representations are visualised, vendor identities and legends are removed for data privacy.

We do not advocate for a fully automated pipeline, and we stress the importance of humans-in-the-loop with professional accounting skills. This material is intended for academic and educational purposes only, and the authors and their institutions disclaim liability for any outcomes arising from its use.  

We partially release the code, removing any identifying information and preserving part of the code to protect the firm's IP.

\bibliography{custom} 

\clearpage

\appendix
\setcounter{figure}{0}
\renewcommand{\thefigure}{A\arabic{figure}}
\setcounter{table}{0}
\renewcommand{\thetable}{A\arabic{table}}

\section{Appendix}\label{sec:appendix}

\subsection{Experimental details}\label{sec:exp_details}
\textbf{Frozen transformers.}

For each model and input type, we freeze the encoder and train a linear classification head on top of the sentence embedding. We use a cross-entropy loss weighted by inverse class frequency, the AdamW optimiser with $5\mathrm{e}{-5}$ learning rate and linear rate, batch size 16, and up to 60 epochs with early stopping on validation weighted F1.The dataset is split 60/20/20 into train, validation and test sets with stratified sampling.
For DeBERTa CLS, we use the model's standard sequence-classification head; for both SBERT and DeBERTa, we use standard mean pooling.

\textbf{LLM baseline.} We implement a baseline for zero-shot classification using Qwen3-4B \citep{qwen3technicalreport}, a small model matching the in-house deployment computation resources constraint. Due to data confidentiality we excluded API-hosted LLMs, which could improve performance. We classify each invoice independently with the following prompt: \emph{You are an expert UK bookkeeper assigning invoices to General Ledger categories. Choose exactly one category from the list. Definitions: \{\}. Invoice: \{\}. Only answer with the name of the category, do not answer with any other text}. The category definitions were provided by the firm (see Table \ref{tab:account_categories}), and the invoice text is as in input B. We disable thinking mode and use greedy decoding for reproducibility. We used a deterministic parser to map the model outputs to a canonical label and count as error outputs that couldn't be parsed to a category. 

\textbf{Fine tuning}
We fine-tune both language models by training the last 2, 4 or 6 layers with the classification head. We use a smaller learning rate of $1\mathrm{e}{-5}$ for the transformer layers than for the classification head ($5\mathrm{e}{-5}$), and optimise the two parameter groups jointly with AdamW and add 10\% linear warm-up which helps DeBERTa to avoid early divergence. Each configuration is run with and without the business-name prefix (+ b.n.). The rest of the parameters are the same. 

\textbf{Computing resources.} Fine-tuning was performed on an NVIDIA RTX PRO 500 Black with 6 GB VRAM, using CUDA 13.2 and driver version 595.71.05. Finetuning SBERT (2 layers + b.n.) took about 10 minutes for 46 epochs, and DeBERTa  (2 layers + b.n.) took approximately 15 minutes for 31 epochs. An inference pass on the 434 test invoices for Qwen3-4B took approximately 5 minutes. 
\onecolumn

\subsection{Supplementary Figures}

\begin{figure}[htb]
    \centering
    \includegraphics[width=0.85\linewidth]{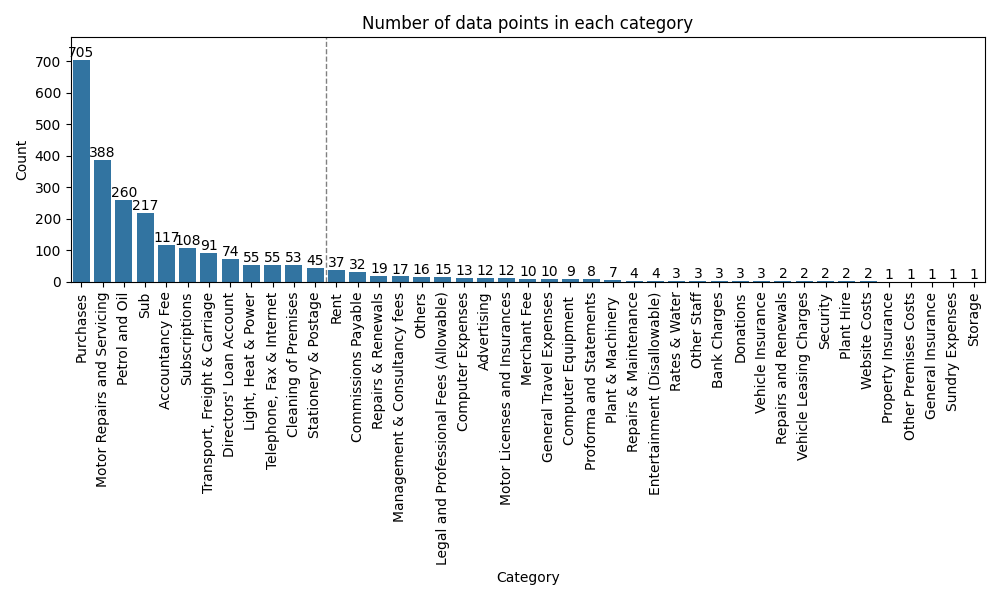}
    \caption{Distribution of invoices across the 42 accounting categories in the full dataset; the categories before the dashed line are included in the analysis.}
    \label{fig:categories_distribution}
\end{figure}

\begin{figure}
    \centering
    \includegraphics[width=1\linewidth]{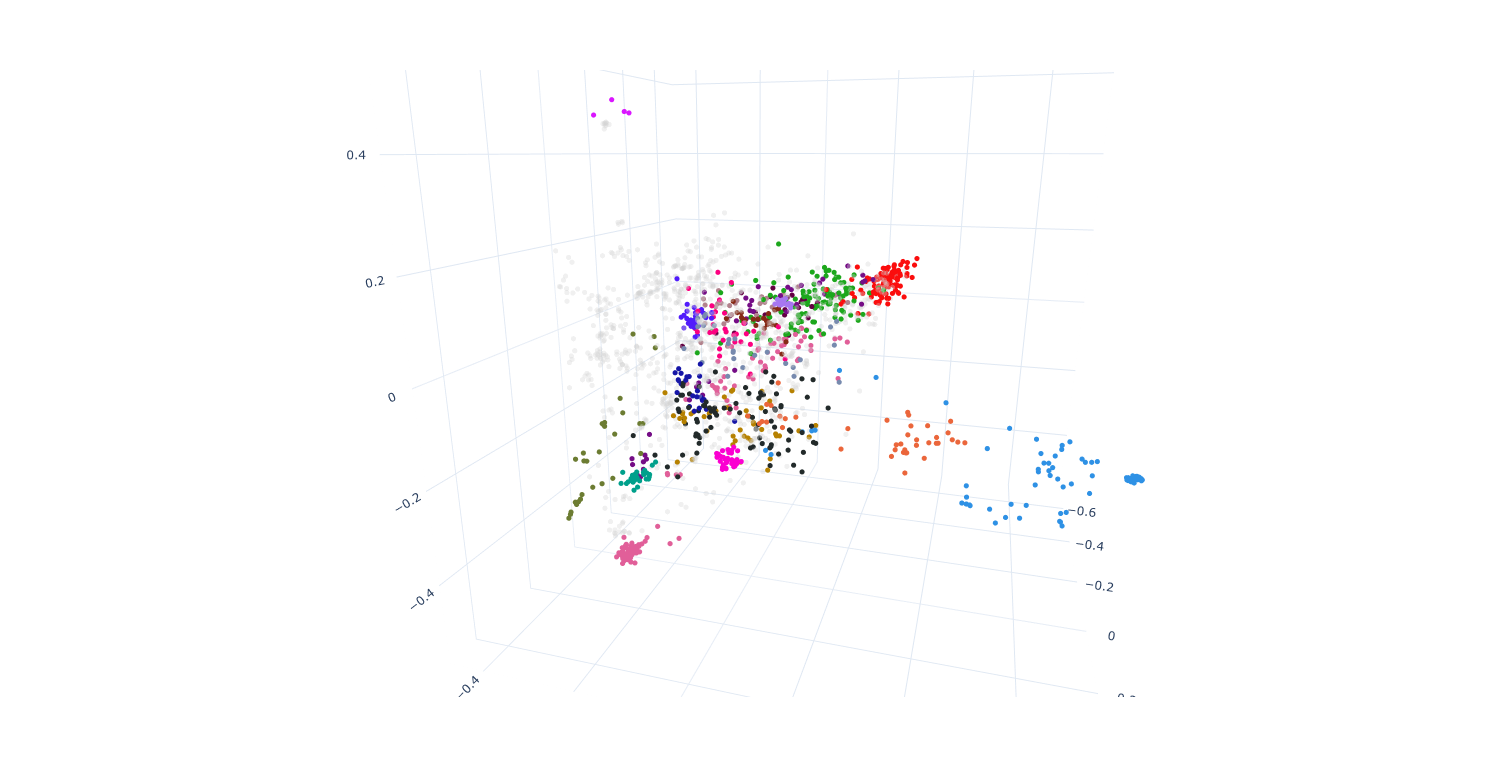}
    \caption{Visualisation of the first three principal components of the representation space of pre-trained SBERT with input B. Points coloured by Vendor name for the 20 most popular vendors, grey otherwise. Legend removed for data privacy. }
    \label{fig:PCA_color_vendors}
\end{figure}

\clearpage

\begin{figure}[h]
    \centering
    \begin{tabular}{cc}
        \includegraphics[width=0.45\textwidth]{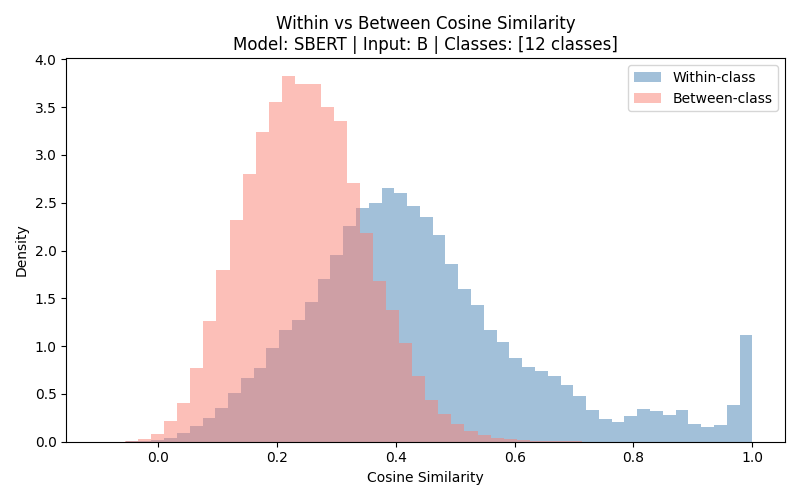} &
        \includegraphics[width=0.45\textwidth]{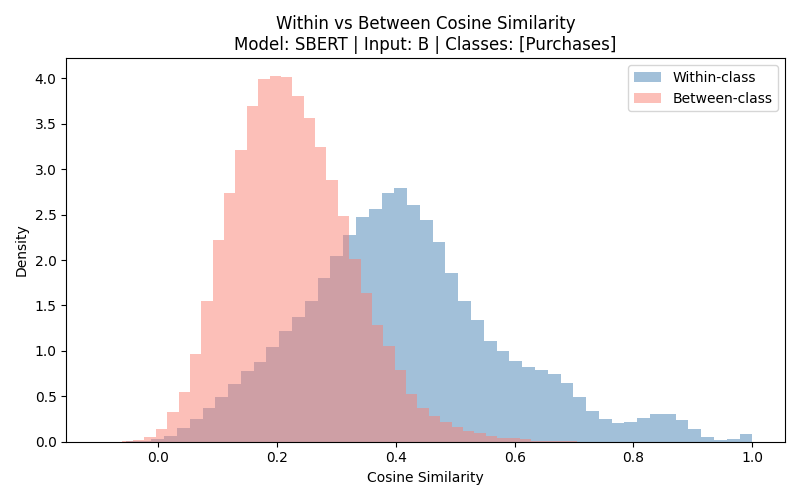} \\
        \includegraphics[width=0.45\textwidth]{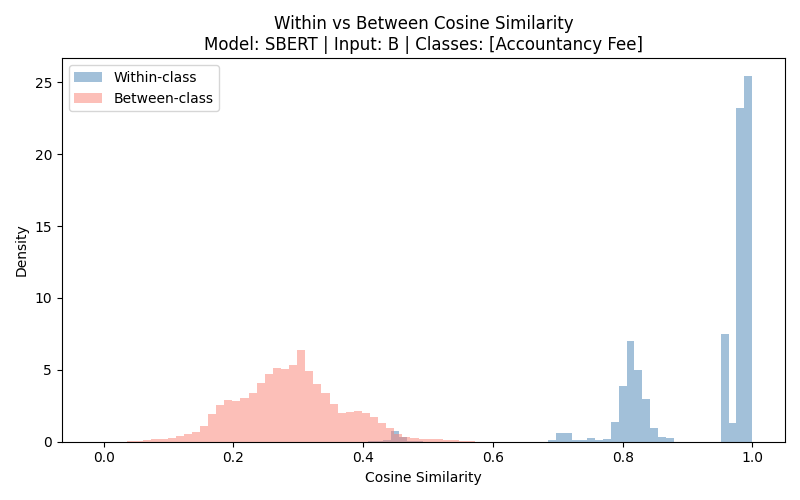} &
        \includegraphics[width=0.45\textwidth]{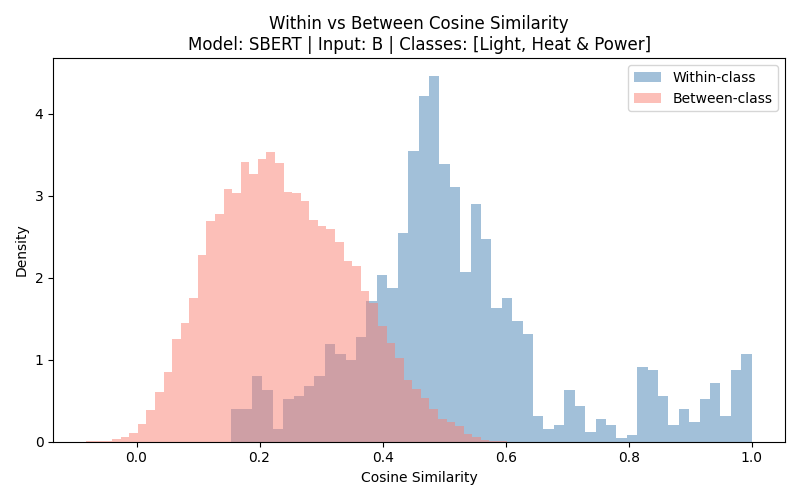} \\
    \end{tabular}
    \caption{Within and between cosine similarity using SBERT Frozen transformer on input B, showing all-class average, Purchases, Accountancy Fee and Light, Heat \& Power. With the exception of Accountancy Fee, they all overlap showing that the direction of the angular variation in embedding space is important for successful classification.}
    \label{fig:cosine_grid}
\end{figure}

\clearpage

\begin{figure}[h]
    \centering
    \includegraphics[width=1\linewidth]{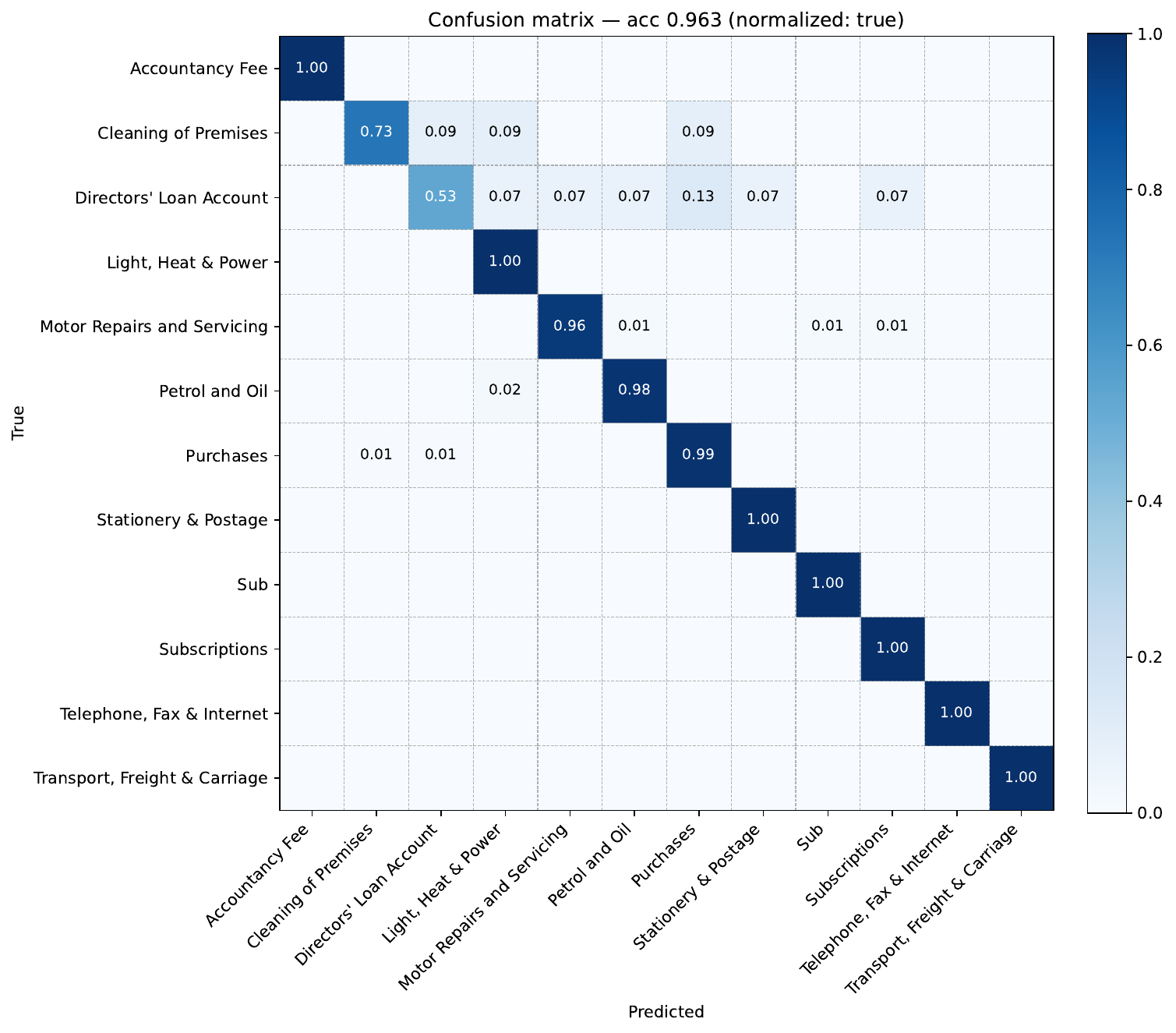}
    \caption{Confusion matrix for SBERT fine-tuned with 2 layers and business name/nature prefix.}
    \label{fig:confusion_matrix}
\end{figure}

\clearpage

\subsection{Supplementary Tables}

\begin{table}[htbp]
\centering
\small
\begin{tabular}{p{4.0cm} p{10.0cm}}
\hline
\textbf{Account Category} & \textbf{Definition} \\
\hline

Purchases &
Cost of goods, raw materials or components bought for resale or for consumption in production during the period. A profit-and-loss item forming part of cost of sales; it records the buying flow, not stock held. Excludes capital assets and overhead services. \\
\hline
Directors' Loan Account &
Related-party control account for amounts a director personally lends to, or withdraws from, the company other than salary, dividend or reimbursed expenses. May be in credit (company owes the director — a liability) or overdrawn (director owes the company — an asset); overdrawn balances attract a s455 CTA 2010 charge and beneficial-loan implications. \\
\hline
Petrol and Oil &
Fuel and lubricant costs incurred in operating business motor vehicles. \\
\hline
Sub &
Abbreviation for Subcontractor. Payments for subcontractor or subcontract labour directly attributable to revenue-generating work. \\
\hline
Motor Repairs and Servicing &
Maintenance, servicing and repair of company vehicles. \\
\hline
Subscriptions &
Recurring professional-body memberships, journals and software or service subscriptions of an administrative nature.\\
\hline
Stationery \& Postage &
Office consumables, printing materials and postal or courier charges.\\
\hline
Cleaning of Premises &
Janitorial costs of keeping premises clean — cleaning staff or contractors, services and cleaning consumables. Excludes repair and upkeep of premises, which is maintenance (see Repairs \& Renewals). \\
\hline

Light, Heat \& Power &
Utility costs — electricity, gas, heating and power — consumed in operating business premises. \\
\hline
Telephone, Fax \& Internet &
Communication costs: telephone, broadband, internet access and related charges\\
\hline
Transport, Freight \& Carriage &
Inwards - Cost of transporting purchased goods or materials into the business. Added to the cost of purchases and therefore part of cost of sales. Outwards - Cost of delivering finished goods out to customers. A distribution/selling overhead, not cost of sales. \\
\hline
Accountancy Fee &
Fees for preparation of statutory and/or management accounts, bookkeeping, tax compliance and related non-audit advisory services. Excludes statutory audit. \\

\hline
\end{tabular}

\caption{Definitions for accounting categories used in the dataset.
There is no standardised UK nominal ledger and nominal ledger code numbers are a convention of the different accounting software. The UK firm that provided our data uses a custom accounting categorisation and proprietary software.
When an invoice contains line items spanning multiple categories, the invoice is assigned to the category with the largest total value, as is standard practice in this UK firm.
}

\label{tab:account_categories}
\end{table}

\clearpage

\clearpage

\begin{table}[h]
\centering
\caption{Average within-class and between-class cosine similarity, using the true labels.}
\label{tab:cosine_wb_true_full}
\centering
\subcaption{Uncentered}

\small
\begin{tabular}{llcc}
\toprule
Type & Model & Within & Between \\
\midrule
A & SBERT        & 0.503 &               0.348 \\
A & DeBERTa CLS     & 0.999&              0.999\\
A & DeBERTa mean   & 0.977  &               0.972 \\
\midrule
B & SBERT        & 0.478 &0.249 \\
B & DeBERTa CLS     & 0.995& 0.993 \\
B & DeBERTa mean   & 0.938 &   0.930\\
\midrule
C & SBERT        &0.641&  0.485\\
C & DeBERTa CLS     & 0.997 & 0.995 \\
C & DeBERTa mean  & 0.956 & 0.948 \\
\bottomrule
\end{tabular}

\centering
\vspace{0.2cm}
\subcaption{Centred}

\small
\begin{tabular}{llcc}
\toprule
Type & Model & Within & Between \\
\midrule
A & SBERT        & -0.005 &  0.000\\
A & DeBERTa CLS    & 0.000 & 0.000 \\
A & DeBERTa mean   &0.000 & 0.002 \\
\midrule
B & SBERT        & 0.003& 0.000\\
B & DeBERTa CLS    & 0.024& 0.016 \\
B & DeBERTa mean    & 0.013&  0.009 \\
\midrule
C & SBERT        & 0.003 & 0.000 \\
C & DeBERTa CLS     & 0.041  &  0.029 \\
C & DeBERTa mean   &  0.019 &  0.011 \\
\bottomrule
\end{tabular}

\end{table}

\begin{table}[h]
\centering
\caption{Average within-cluster and between-cluster cosine similarity.}
\label{tab:cosine_wb_cluster_full}
\centering
\subcaption{Uncentered}

\small
\begin{tabular}{llcc}
\toprule
Type & Model & Within & Between \\
\midrule
A & SBERT        & 0.820  &              0.341 \\
A & DeBERTa CLS     & 0.999& 0.999\\
A & DeBERTa mean   & 0.991 & 0.970 \\
\midrule
B & SBERT        & 0.708&0.265\\
B & DeBERTa CLS     & 0.997&            0.981 \\
B & DeBERTa mean   & 0.942&0.841\\
\midrule
C & SBERT        &0.823&0.476\\
C & DeBERTa CLS     & 0.998&0.986\\
C & DeBERTa mean  & 0.953 & 0.864\\
\bottomrule
\end{tabular}

\centering
\subcaption{Centred}

\small
\begin{tabular}{llcc}
\toprule
Type & Model & Within & Between \\
\midrule
A & SBERT        & -0.005 &  0.000\\
A & DeBERTa CLS    & 0.000 & 0.000 \\
A & DeBERTa mean   &0.000 & 0.002 \\
\midrule
B & SBERT        & 0.003& 0.000\\
B & DeBERTa CLS    & 0.024& 0.016 \\
B & DeBERTa mean    & 0.013&  0.009 \\
\midrule
C & SBERT        & 0.003 & 0.000 \\
C & DeBERTa CLS     & 0.041  &  0.029 \\
C & DeBERTa mean   &  0.019 &  0.011 \\
\bottomrule
\end{tabular}

\end{table}

\begin{table}[h]
\setlength{\tabcolsep}{3pt}
\centering
\caption{Classification Report (precision, recall, F1) - SBERT. We report the mean$\pm$s.d. over 5 splits.}\label{tab:report_SBERTB}
\small
\begin{tabular}{l ccc | ccc |ccc }
\toprule
 & \multicolumn{3}{c|}{\textbf{Input A}} & \multicolumn{3}{c|}{\textbf{Input B}} & \multicolumn{3}{c}{\textbf{Input C}} \\
\cmidrule(lr){2-4} \cmidrule(lr){5-7} \cmidrule(lr){8-10}
\textbf{Class} & \textbf{P} & \textbf{R} & \textbf{F1} & \textbf{P} & \textbf{R} & \textbf{F1} & \textbf{P} & \textbf{R} & \textbf{F1} \\
\midrule
Accountancy Fee               & $0.91_{\pm0.02}$	&$0.99_{\pm0.02}$	&$0.95_{\pm0.01}$ & $0.79_{\pm0.06}$&	$0.99_{\pm0.02}$	&$0.88_{\pm0.04}$&$0.64_{\pm0.06}$	&$0.99_{\pm0.02}$	&$0.78_{\pm0.05}$  \\
Cleaning of Premises          & $ 0.74_{\pm0.09}$&	$0.55_{\pm0.14}$	&$0.63_{\pm0.13}$  & $0.92_{\pm0.07}$	&$0.65_{\pm0.08}$	&$0.76_{\pm0.08}$ & $0.93_{\pm0.10}$	&$0.58_{\pm0.15}$&	$0.71_{\pm0.14}$   \\
Directors' Loan Account       & $0.48_{\pm0.38}$	&$0.11_{\pm0.08}$	&$0.17_{\pm0.13}$ & $0.53_{\pm0.36}	$&$0.07_{\pm0.05}$&	$0.12_{\pm0.08 }$&$ 0.50_{\pm0.50}$	&$0.00_{\pm0.06}$&	$0.10_{\pm0.10}$  \\
Light, Heat \& Power          &  $0.86_{\pm0.13}$	&$0.71_{\pm0.24}$&	$0.76_{\pm0.20}$& $0.94_{\pm0.07}$	&$0.96_{\pm0.05}$&	$0.95_{\pm0.02}$ & $0.95_{\pm0.07}$&	$0.95_{\pm0.08}$	&$0.95_{\pm0.04}  $  \\
Motor Repairs and Servicing & $0.77_{\pm0.04}$&	$0.87_{\pm 0.04}$	&$0.82_{\pm0.04}$  & $0.90_{\pm0.06}$	&$0.91_{\pm0.03}$	&$0.90_{\pm0.03}$ &$0.92_{\pm0.02}$	&$0.87_{\pm0.02}$&	$0.90_{\pm0.02}$\\
Petrol and Oil                & $0.47_{\pm0.02}$	&$0.81_{\pm0.04}$	&$0.60_{\pm0.03}$& $0.81_{\pm0.07}	$&$0.94_{\pm0.03}$	&$0.87_{\pm0.04}$& $0.74_{\pm0.06}$	&$0.92_{\pm0.02}$&	$0.82_{\pm0.04}$  \\
Purchases                     &$ 0.86_{\pm0.04}	$&$0.53_{\pm0.04}	$&$0.66_{\pm0.04} $& $0.90_{\pm0.02}$&	$0.89_{\pm 0.03}$	&$0.89_{\pm0.02}$ & $0.90_{\pm0.03}$	&$0.83_{\pm0.04}$	&$0.86_{\pm0.03}$  \\

Stationery \& Postage&  $0.81_{\pm0.14}$	&$0.69_{\pm0.09}$&	$0.74_{\pm0.10}$  &  $0.91_{\pm0.13}$&	$0.82_{\pm0.10}$&	$0.85_{\pm0.06}$ &$0.94_{\pm0.13}$	&$0.71_{\pm0.10}$	&$0.80_{\pm0.08}$\\

Sub    &   $0.80_{\pm0.04}$	&$0.70_{\pm0.07}$&	$0.75_{\pm0.05}$ &   $0.93_{\pm0.01}$&	$0.87_{\pm0.05}$	&$0.90_{\pm0.03}$   &$0.73_{\pm0.07}$&	$0.94_{\pm0.03}$	&$0.82_{\pm0.03}$\\

Subscriptions                 & $0.59_{\pm0.06}$	&$0.85_{\pm0.11}$	&$0.70_{\pm0.08}$ &$0.81_{\pm0.03}$& $0.82_{\pm0.09}	$&$0.81_{\pm0.04}$ & $0.90_{\pm0.04}$&	$0.75_{\pm0.09}$	&$0.81_{\pm0.07} $  \\
Telephone, Fax \& Internet    &$0.37_{\pm0.11}$&	$0.73_{\pm0.17}$	&$0.49_{\pm0.13} $    & $0.76_{\pm 0.11}$&	$1.00_{\pm0.00}$	&$0.86_{\pm0.07}$& $0.75_{\pm0.08}$	&$0.96_{\pm0.05}$&	$0.84_{\pm0.05 }$  \\
Transport, Freight \& Carriage & $0.60_{\pm0.11}$	&$0.90_{\pm0.10}$	&$0.72_{\pm0.11} $&$ 0.83_{\pm0.07}	$&$0.89_{\pm0.09}$&	$0.86_{\pm0.06} $     &$ 0.94_{\pm0.08}$	&$0.77_{\pm0.11}$&	$0.84_{\pm0.07} $  \\

\bottomrule
\end{tabular}
\end{table}

\begin{table}[h]
\centering
\setlength{\tabcolsep}{3pt}
\caption{Classification Report (precision, recall, F1) - DeBERTa CLS. We report the mean$\pm$s.d. over 5 splits.}
\small
\begin{tabular}{l ccc | ccc |ccc }
\toprule
 & \multicolumn{3}{c|}{\textbf{Input A}} & \multicolumn{3}{c|}{\textbf{Input B}} & \multicolumn{3}{c}{\textbf{Input C}}  \\
\cmidrule(lr){2-4} \cmidrule(lr){5-7} \cmidrule(lr){8-10}
\textbf{Class} & \textbf{P} & \textbf{R} & \textbf{F1} & \textbf{P} & \textbf{R} & \textbf{F1} & \textbf{P} & \textbf{R} & \textbf{F1}  \\
\midrule
Accountancy Fee               & $0.15_{\pm0.15}$&	$0.59_{\pm0.54}$	&$0.23_{\pm0.23}$  & $0.81_{\pm0.08}$	&$0.97_{\pm 0.02}$	&$0.88_{\pm 0.05} $      &$0.61_{\pm0.24}$&	$0.83_{\pm0.04	}$&$0.68_{\pm0.17 }$ \\
Cleaning of Premises          & $0.00_{\pm0.00}$	&$0.00 _{\pm0.00}$	&$0.00_{\pm0.00}$& $0.56_{\pm0.27}$	&$0.80_{\pm0.04}$	&$0.62_{\pm0.17}$& $0.31_{\pm 0.19}$	&$0.71_{\pm 0.22}$&	$0.41_{\pm 0.19 }$\\
Directors' Loan Account       & $0.01_{\pm0.02}$	&$0.03_{\pm0.06}$	&$0.01_{\pm0.03}$ & $0.15_{\pm 0.29}$	&$0.04_{\pm0.06}$	&$0.06_{\pm0.10} $& $0.20_{\pm 0.45}$&	$0.01_{\pm0.03}	$&$0.03_{\pm0.06}$  \\
Light, Heat \& Power          & $0.22_{\pm0.44}$	&$0.16_{\pm0.28}$	&$0.10_{\pm 0.14}$ &   $0.57_{\pm 0.21}$&	$0.65_{\pm 0.10	}$&$0.60_{\pm0.15   } $ & $0.10_{\pm0.19}$	&$0.20_{\pm0.28}$&	$0.13_{\pm 0.21}$   \\
Motor Repairs and Servicing &  $0.52_{\pm0.09}$&	$0.65_{\pm0.14}$&	$0.56_{\pm0.05}$ &  $0.78_{\pm 0.05	}$&$0.61_{\pm0.04}$&	$0.68_{\pm 0.03 }$&$0.51_{\pm0.04}$	&$0.55_{\pm0.06}$&	$0.53_{\pm0.03}$ \\
Petrol and Oil & $0.05_{\pm0.11}$&	$0.06_{\pm0.14}$	&$0.05_{\pm0.12}$ &   $0.52_{\pm0.14}$	&$0.83_{\pm0.08}$&	$0.62_{\pm 0.06 }$ &   $0.44_{\pm0.10}$&	$0.67_{\pm0.17}$	&$0.51_{\pm0.07}$ \\
Purchases & $0.52_{\pm0.14}$&	$0.55_{\pm0.21}$&	$0.49_{\pm0.06}$ & $0.86_{\pm0.03}$	&$0.55_{\pm0.09	}$&$0.67_{\pm0.06}$&   $0.84_{\pm0.06	}$&$0.43_{\pm0.07}$&	$0.57_{\pm0.06}$  \\
Stationery \& Postage  & $0.16_{\pm0.23}$&	$0.13_{\pm0.20}	$&$0.13_{\pm0.18}$&  $0.44_{\pm0.12}$&	$0.29_{\pm0.13}$	&$0.33_{\pm 0.11}$  &$0.05_{\pm0.07}$	&$0.07_{\pm0.10}$	&$0.06_{\pm 0.08}$ \\
Sub   &$0.29_{\pm0.29}$	&$0.30_{\pm0.28}$&	$0.29_{\pm0.27}$&   $0.60_{\pm0.07}$	&$0.80_{\pm0.07}$&	$0.68_{\pm0.05 }$&$0.53_{\pm0.10}	$&$0.70_{\pm0.14}$	&$0.59_{\pm0.05}$\\
Subscriptions & $0.08_{\pm0.12	}$&$0.07_{\pm0.14}$&	$0.07_{\pm0.13}$ & $0.44_{\pm0.10}$	&$0.55_{\pm 0.23}$	&$0.46_{\pm0.10} $   & $0.37_{\pm0.18}$	&$0.25_{\pm 0.22}$	&$0.26_{\pm0.16} $\\
Telephone, Fax \& Internet    & $0.42_{\pm0.53}$&	$0.42_{\pm0.40}$&	$0.34_{\pm0.38}$&$ 0.69_{\pm0.20}$	&$0.67_{\pm 0.05}$	&$0.66_{\pm0.11}$& $0.33_{\pm0.24}$	&$0.27_{\pm0.22}$	&$0.26_{\pm0.16}$ \\
Transport, Freight \& Carriage & $0.00_{\pm0.00}$	&$0.00_{\pm0.00}$&	$0.00_{\pm0.00}$ & $0.35_{\pm0.02}$&	$0.40_{\pm0.18}$&	$0.36_{\pm0.10}$ & $0.21_{\pm0.19	}$&$0.12_{\pm0.08}$&	$0.14_{\pm0.09}$ \\

\bottomrule
\end{tabular}
\end{table}

\begin{table}[h]
\setlength{\tabcolsep}{3pt}
\centering
\caption{Classification Report (precision, recall, F1) - DeBERTa mean. We report the mean$\pm$s.d. over 5 splits.}
\small
\begin{tabular}{l ccc |ccc |ccc}
\toprule
 & \multicolumn{3}{c|}{\textbf{Input A}} & \multicolumn{3}{c|}{\textbf{Input B}} & \multicolumn{3}{c}{\textbf{Input C}} \\
\cmidrule(lr){2-4} \cmidrule(lr){5-7} \cmidrule(lr){8-10}
\textbf{Class} & \textbf{P} & \textbf{R} & \textbf{F1} & \textbf{P} & \textbf{R} & \textbf{F1} & \textbf{P} & \textbf{R} & \textbf{F1}  \\
\midrule
Accountancy Fee &$0.46_{\pm0.07}$&	$1.00_{\pm0.00}$&	$0.63_{\pm0.07}$ & $0.34_{\pm0.05}$	&$0.99_{\pm0.02}	$&$0.51_{\pm0.05}$  &  $0.37_{\pm0.06}$&	$0.92_{\pm0.10}$	&$0.52_{\pm0.07}$  \\
Cleaning of Premises          & $0.00_{\pm0.00}$&	$0.00_{\pm0.00}$&	$0.00_{\pm0.00}$&   $0.74_{\pm0.06}$	&$0.36_{\pm0.09}	$&$0.48_{\pm0.09}$    &  $0.85_{\pm0.22}$&	$0.29_{\pm0.08}$&	$0.43_{\pm0.11}$ \\
Directors' Loan Account       &   $0.22_{\pm0.22}$	& $0.04_{\pm0.04}$&	$0.07_{\pm0.06}$ & $0.23_{\pm0.20}$&	$0.08_{\pm0.06}$	&$0.11_{\pm0.08}$ & $0.07_{\pm0.09}$	&$0.03_{\pm0.04}$	&$0.04_{\pm0.05}$   \\
Light, Heat \& Power          &  $0.11_{\pm0.06}$&	$0.18_{\pm0.11}$	&$0.12_{\pm0.05} $& $0.55_{\pm0.17}$&	$0.47_{\pm0.24}$	&$0.50_{\pm0.21}$  & $0.43_{\pm0.08}$&	$0.58_{\pm0.22}$	&$0.49_{\pm0.13}$  \\
Motor Repairs and Servicing  & $0.62_{\pm0.07}$	&$0.72_{\pm0.08}$	&$0.66_{\pm0.03}$&     $0.73_{\pm0.02}$&	$0.69_{\pm 0.07}	$&$0.71_{\pm0.04} $&$0.69_{\pm0.05}$	&$0.68_{\pm0.07}$	&$0.68_{\pm0.03}$\\
Petrol and Oil                & $0.36_{\pm0.04}$&	$0.46_{\pm0.05}$	&$0.40_{\pm0.04}$	 &$ 0.65_{\pm0.06}$	&$0.77_{\pm0.05}$	&$0.70_{\pm0.03}$       & $0.72_{\pm0.05}$	&$0.74_{\pm0.05}$&	$0.73_{\pm0.05}$   \\
Purchases                     &$0.60_{\pm0.06}$&	$0.51_{\pm0.08}$	&$0.55_{\pm0.04}$& $0.81_{\pm0.04}$&	$0.64_{\pm0.05}$	&$0.71_{\pm0.04}$    & $0.80_{\pm0.02}$	&$0.65_{\pm0.05}$	&$0.72_{\pm0.03}$  \\
Stationery \& Postage   &$0.80_{\pm0.45}$&	$0.20_{\pm0.16}$	&$0.31_{\pm0.23}$&   $0.78_{\pm0.22}$	&$0.33_{\pm0.21}$	&$0.43_{\pm0.19}$    &$0.56_{\pm0.29}$&	$0.20_{\pm0.09}$	&$0.29_{\pm0.14}$\\

Sub    &$0.74_{\pm0.08}$&	$0.43_{\pm0.05}$	&$0.54_{\pm0.03}$ &  $0.85_{\pm0.04}$	&$0.73_{\pm0.06}$	&$0.78_{\pm0.03}$  & $0.56_{\pm0.04}$	&$0.76_{\pm0.05}$	&$0.64_{\pm0.04}$\\
Subscriptions                 & $0.40_{\pm0.11}$	&$0.49_{\pm0.10}$&	$0.43_{\pm0.06}$ &$0.41_{\pm0.09}$	&$0.55_{\pm0.06}$	&$0.46_{\pm0.08}$     & $0.55_{\pm0.09}$	&$0.49_{\pm0.09}$&	$0.52_{\pm0.08}$ \\
Telephone, Fax \& Internet    &  $0.58_{\pm0.26}$	&$0.53_{\pm0.15}$	&$0.52_{\pm0.14}$& $0.67_{\pm0.16}$	&$0.67_{\pm0.21}$	&$0.66_{\pm0.16}$& $0.63_{\pm0.18}$&	$0.64_{\pm0.21}$	&$0.63_{\pm0.18}$  \\
Transport, Freight \& Carriage & $0.59_{\pm0.14}$&	$0.28_{\pm0.08}$	&$0.37_{\pm0.09}$ &$0.31_{\pm0.16}$&	$0.26_{\pm0.12}$	&$0.28_{\pm0.14}$ & $0.26_{\pm0.10}$&	$0.21_{\pm0.09}$	&$0.23_{\pm0.09} $    \\

\bottomrule
\end{tabular}
\end{table}

\begin{table}[htbp]
\centering
\begin{tabular}{lrrrr}
\toprule
 & Precision & Recall & F1-score  \\
\midrule
Accountancy Fee               &$ 0.22_{\pm0.05}$&  $0.23_{\pm0.05}$&$ 0.23_{\pm0.05}$\\
Cleaning of Premises          & $0.68_{\pm0.22}$ & $0.33_{\pm0.08}$&$0.43_{\pm0.10}$ \\
Directors' Loan Account       & $0.00_{\pm0.00}$    & $0.00_{\pm0.00}$   &     $0.00_{\pm0.00}$    \\
Light, Heat \& Power          & $0.29_{\pm0.02} $ &$1.00_{\pm 0.00}$&     $0.45_{\pm0.03}$ \\
Motor Repairs and Servicing   & $0.58_{\pm0.08}$  &$0.36_{\pm0.05}$&    $  0.44_{\pm0.06}$  \\
Petrol and Oil                & $0.93_{\pm0.04} $ &$0.62_{\pm0.05}$&   $  0.75_{\pm0.04}$\\
Purchases                     & $0.79_{\pm0.01} $ &$0.85_{\pm0.03}$& $ 0.82_{\pm0.02}$\\
Stationery \& Postage         & $0.67_{\pm0.20}$  &$0.29_{\pm0.17}$&  $ 0.37_{\pm0.17}$   \\
Sub                           & $0.47_{\pm0.45}$  &$0.03_{\pm0.03}$&  $0.05_{\pm0.05}$ \\
Subscriptions                 &$0.17_{\pm0.02}$ &$0.46_{\pm0.09} $ & $0.25_{\pm0.04}$\\
Telephone, Fax \& Internet    & $0.82_{\pm0.17}$ & $0.35_{\pm0.04}$&  $0.48_{\pm0.06}$\\
Transport, Freight \& Carriage &$ 0.30_{\pm0.02} $& $0.50_{\pm{0.07}}$ &  $0.37_{\pm0.03}$\\

\bottomrule
\end{tabular}
\caption{Classification report for the LLM baseline, mean$\pm$s.d. over 5 splits.}
\label{tab:class_rep_baseline_LLM}
\end{table}

\begin{table}[htbp]
\centering
\begin{tabular}{lrrr}
\toprule
 & Precision & Recall & F1-score \\
\midrule
Accountancy Fee               & $1.00_{\pm0.00}$  &$0.99_{\pm0.02}$ & $1.00_{\pm0.01}$  \\
Cleaning of Premises          & $0.88_{\pm0.16} $& $0.75_{\pm0.08}$   & $0.80_{\pm0.09}$  \\
Directors' Loan Account       & $0.96_{\pm0.09}$&  $0.29_{\pm0.12}$  & $0.44_{\pm0.15}$  \\
Light, Heat \& Power          & $0.95_{\pm0.05} $& $0.98_{\pm0.04}$& $0.96_{\pm0.02}$ \\
Motor Repairs and Servicing   &  $0.99_{\pm0.01}$&  $0.92_{\pm0.02} $   & $0.95_{\pm0.01}$ \\
Petrol and Oil                &  $0.77_{\pm0.04}$ & $0.88_{\pm0.03}$   & $0.82_{\pm0.03} $     \\
Purchases                     & $0.82_{\pm0.02}$&  $0.93_{\pm0.02}$ & $0.87_{\pm0.01}$ \\
Stationery \& Postage         & $0.93_{\pm0.06}$&  $0.82_{\pm0.17}$& $0.86_{\pm0.10}$ \\
Sub                           & $1.00_{\pm0.00}$& $0.90_{\pm0.04}$& $0.95_{\pm0.02}$ \\
Subscriptions                 & $0.98_{\pm0.03}$&  $0.91_{\pm0.09}$ &$0.94_{\pm0.05} $\\
Telephone, Fax \& Internet    &$1.00_{\pm0.00}$&   $0.91_{\pm0.09}$   & $0.95_{\pm0.05} $    \\
Transport, Freight \& Carriage & $1.00_{\pm0.00}$&  $0.99_{\pm0.02}$ & $0.99_{\pm0.01}$ \\
\bottomrule
\end{tabular}
\caption{Classification report for the vendor baseline. We report the mean$\pm$s.d. over 5 splits.}
\label{tab:class_rep_baseline}
\end{table}

\FloatBarrier

\clearpage

\begin{table}[h]
\centering
\small
\begin{tabular}{lccc}
\hline
\textbf{Model}  & \textbf{Acc.} & \textbf{M F1} & \textbf{W F1} \\
\hline
SBERT - 2 l & 0.880 & 0.845 &  0.882\\
SBERT - 4 l &  0.876 & 0.845 & 0.874\\
SBERT - 6 l & 0.917 & \textbf{0.875} & 0.914\\
\hline
SBERT - 2 l + b.n. & 0.903 & 0.839 & 0.904\\
SBERT - 4 l + b.n. & \textbf{0.931} & 0.874 & 0.929\\
SBERT - 6 l + b.n. & 0.929 & 0.873 & \textbf{0.927}\\
\hline
DeBERTa - 2 l & 0.846 & 0.774 & 0.842\\
DeBERTa - 4 l & 0.664 & 0.471 & 0.651\\
DeBERTa - 6 l &  0.433 & 0.309 & 0.411\\
\hline
DeBERTa - 2 l + b.n. & 0.839 & 0.766 & 0.843\\
DeBERTa - 4 l + b.n.  &  0.406 & 0.108 & 0.302 \\
DeBERTa - 6 l + b.n. & 0.325 & 0.041 & 0.159 \\
\hline
DeBERTa mean - 2 l& 0.869 & 0.801& 0.863 \\
DeBERTa mean - 4 l & 0.634 & 0.442 & 0.601\\
DeBERTa mean - 6 l & 0.325 & 0.041 & 0.159 \\
\hline
DeBERTa mean - 2 l + b.n. &  0.922 & 0.845 & 0.915\\
DeBERTa mean - 4 l + b.n. & 0.684 & 0.436 & 0.664\\
DeBERTa mean - 6 l + b.n. &  0.442 & 0.143 & 0.396 \\
\hline
\end{tabular}
\caption{Accuracy, macro F1 and weighted F1 across
fine-tuning configurations and number of unfrozen final
layers (l) for input type A with and without business-
nature prefix (+b.n.), for a single split. The best performance is in \textbf{bold}.}
\label{tab:class_ft_full}
\end{table}

\begin{table}[h]
\centering
\small
\begin{tabular}{lccc}
\hline
\textbf{Model}  & \textbf{Acc.} & \textbf{M F1} & \textbf{W F1} \\
\hline
SBERT - 2 l & 0.942 & 0.892 & 0.939\\
SBERT - 4 l &  0.945 & 0.894 & 0.938\\
SBERT - 6 l & 0.938 & 0.898 & 0.934\\
\hline
SBERT - 2 l + b.n. &  0.940 & 0.895 & 0.937\\
SBERT - 4 l + b.n. & \textbf{0.959} & 0.923 & \textbf{0.956}\\
SBERT - 6 l + b.n. & \textbf{0.959} & \textbf{0.931} & \textbf{0.956}\\
\hline
DeBERTa - 2 l & 0.878 & 0.776 & 0.872\\
DeBERTa - 4 l & 0.535 & 0.247 & 0.494\\
DeBERTa - 6 l & 0.325 & 0.041 & 0.159 \\
\hline
DeBERTa - 2 l + b.n. & 0.887 & 0.812 & 0.884\\
DeBERTa - 4 l + b.n.  & 0.537 & 0.181 & 0.448\\
DeBERTa - 6 l + b.n. & 0.325 & 0.041 & 0.159 \\
\hline
DeBERTa mean - 2 l& 0.917 & 0.861 & 0.914\\
DeBERTa mean - 4 l & 0.781 & 0.671 & 0.783\\
DeBERTa mean - 6 l & 0.353 & 0.073 & 0.214\\
\hline
DeBERTa mean - 2 l + b.n. & 0.933 & 0.886 & 0.930\\
DeBERTa mean - 4 l + b.n. & 0.793 & 0.551 & 0.772\\
DeBERTa mean - 6 l + b.n. & 0.385 & 0.155 & 0.261 \\
\hline
\end{tabular}
\caption{Accuracy, macro F1 and weighted F1 across
fine-tuning configurations and number of unfrozen final
layers (l) for input type C with and without business-
nature prefix (+b.n.), for a single split. The best performance is in \textbf{bold}.}

\end{table}

\begin{table}[h]
\centering
\caption{Classification Report (precision, recall, F1) - SBERT fine-tuned with 2 unfrozen layers + business name prefix. We report the mean$\pm$s.d. over 5 splits.}
\label{tab:class_un1}
\begin{tabular}{l ccc  }
\toprule
 & \textbf{Precision} & \textbf{Recall} & \textbf{F1}\\
\midrule
Accountancy Fee    &  $ 1.00_{\pm0.00}$	&$0.99_{\pm0.02}$	&$1.00_{\pm0.01}$\\

Cleaning of Premises & $0.83_{\pm0.09}$	&$0.91_{\pm0.11}$	&$0.86_{\pm0.08}$\\

Directors' Loan Account & $0.86_{\pm0.05}$	&$0.61_{\pm0.12}$	&$0.71_{\pm0.07}$\\

Light, Heat \& Power & $0.92_{\pm0.09}$&	$0.98_{\pm0.04}$	&$0.95_{\pm0.04}$\\

Motor Repairs and Servicing & $0.97_{\pm0.02}$&	$0.96_{\pm0.01}$	&$0.97_{\pm0.01}$\\

Petrol and Oil &$ 0.96_{\pm0.02}$	&$0.97_{\pm0.02}$	&$0.96_{\pm0.01}$\\

Purchases & $0.97_{\pm0.00}$	&$0.98_{\pm 0.01}$	&$0.98_{\pm0.01}$\\

Stationery \& Postage &$0.94_{\pm0.08}$	&$0.98_{\pm0.05}$	&$0.96_{\pm0.04}$\\

Sub & $0.96_{\pm0.02}$	&$0.98_{\pm0.02}$&	$0.97_{\pm0.02}$\\

Subscriptions & $0.97_{\pm 0.04}	$&$0.98_{\pm0.02}$&	$0.98_{\pm0.02}$\\

Telephone, Fax \& Internet & $0.95_{\pm0.07}$&	$1.00_{\pm0.00}$	&$0.97_{\pm0.04}$\\

Transport, Freight \& Carriage & $0.96_{\pm0.07}$&	$0.99_{\pm0.02}$	&$0.97_{\pm 0.05}$\\

\bottomrule
\end{tabular}
\end{table}

\begin{table}[h]
\centering
\caption{Classification Report (precision, recall, F1) - SBERT fine-tuned with 6 unfrozen layers. We report the mean$\pm$s.d. over 5 splits.}
\label{tab:class_un2}
\begin{tabular}{l ccc  }
\toprule
 & \textbf{Precision} & \textbf{Recall} & \textbf{F1} \\
\midrule
Accountancy Fee    &   $1.00_{\pm0.00}$	&$0.99_{\pm0.02}$	&$1.00_{\pm0.01}$\\
 
Cleaning of Premises &  $0.83_{\pm0.13}$&$0.91_{\pm0.11}	$&$0.86_{\pm0.08}$\\

Directors' Loan Account &  $0.71_{\pm0.11}$	&$0.44_{\pm0.14}$	&$0.53_{\pm0.08}$\\

Light, Heat \& Power & $0.94_{\pm0.07}$&$1.00_{\pm 0.00}$&	$0.97_{\pm 0.04}$\\

Motor Repairs and Servicing & $0.99_{\pm0.01}$	&$0.95 _{\pm0.00}$	&$0.97_{\pm 0.01}$ \\

Petrol and Oil &  $0.96_{\pm 0.01}$	&$0.97_{\pm0.02}$	&$0.97_{\pm0.01 }$ \\

Purchases & $0.94_{\pm0.02}$	&$0.97_{\pm0.03}$	&$0.95 _{\pm0.01}$\\

Stationery \& Postage &  $0.94_{\pm0.08}$&	$0.98_{\pm 0.05	}$&$0.96_{\pm0.04}$\\

Sub &$ 0.96_{\pm0.01}$	&$0.99_{\pm0.01}$	&$0.97_{\pm0.01}$\\

Subscriptions & $0.98_{\pm0.02}$	&$0.96_{\pm0.04}$	&$0.97_{\pm0.03}$ \\

Telephone, Fax \& Internet & $0.97_{\pm0.05}$&	$1.00_{\pm0.00}$	&$0.98_{\pm0.02}$\\

Transport, Freight \& Carriage &$ 0.96_{\pm 0.04}$	&$0.99_{\pm 0.02}$&	$0.97_{\pm 0.02}$ \\

\bottomrule
\end{tabular}
\end{table}

\begin{table}[h]
\centering
\caption{Classification Report (precision, recall, F1) - DeBERTa fine-tuned with 2 unfrozen layers + business name prefix. We report the mean$\pm$s.d. over 5 splits. }
\label{tab:class_un3}
\begin{tabular}{l ccc  }
\toprule
 & \textbf{Precision} & \textbf{Recall} & \textbf{F1}\\
\midrule
Accountancy Fee    &   $0.98_{\pm0.04}$&	$0.99_{\pm0.02}$	&$0.99_{\pm0.02}$\\

Cleaning of Premises & $0.98_{\pm0.04}$&	$0.99_{\pm0.02}$&	$0.99_{\pm0.02}$\\

Directors' Loan Account & $0.77_{\pm0.06}$&	$0.64_{\pm 0.2}$	&$0.68_{\pm0.09}$ \\

Light, Heat \& Power &  $1.00_{\pm0.00}$&	$0.98_{\pm0.04}$	&$0.99_{\pm0.02}$ \\

Motor Repairs and Servicing &  $0.98_{\pm0.01}$	&$0.95_{\pm0.02}$	&$0.97_{\pm0.01}$\\

Petrol and Oil & $0.98_{\pm0.03}$&	$0.97_{\pm0.02}$	&$0.98_{\pm0.02}$\\

Purchases & $0.96_{\pm0.01}$&	$0.99_{\pm0.01}$	&$0.97_{\pm0.01}$\\

Stationery \& Postage & $0.86_{\pm0.10}$&	$0.80_{\pm0.18}$&	$0.82_{\pm0.13}$\\

Sub &  $0.96_{\pm0.05}$	&$0.95_{\pm0.03}$&	$0.95_{\pm0.02}$\\

Subscriptions &   $0.93_{\pm0.04}$	&$0.95_{\pm0.05}$	&$0.94_{\pm0.02}$\\

    Telephone, Fax \& Internet &$ 0.96_{\pm0.10}$	&$0.96_{\pm0.05}$	&$0.96_{\pm0.05}$\\

Transport, Freight \& Carriage & $0.94_{\pm0.08}$	&$1.00_{\pm0.00}$&	$0.97_{\pm0.05}$\\

\bottomrule
\end{tabular}
\end{table}

\end{document}